\documentclass[10pt,twocolumn,letterpaper]{article}

\usepackage[pagenumbers]{cvpr} 

\definecolor{cvprblue}{rgb}{0.21,0.49,0.74}
\usepackage[pagebackref,breaklinks,colorlinks,allcolors=cvprblue]{hyperref}

\usepackage{algorithm} 
\usepackage{algorithmic}
\usepackage{booktabs}
\usepackage{orcidlink}
\usepackage{tikz}

\usepackage{multirow}
\usepackage{graphicx}

\usepackage{colortbl}

\usepackage{caption}  
\usepackage{tabularx} 

\usepackage{xcolor}
\newcommand\crule[3][black]{\textcolor{#1}{\rule{#2}{#3}}}

\usepackage{pifont}
\usepackage[normalem]{ulem}
\newcommand{\cmark}{{\color{green}\ding{51}}}
\newcommand{\xmark}{{\color{red}\ding{55}}}

\usepackage[normalem]{ulem} 

\def\paperID{*****} 
\def\confName{CVPR}
\def\confYear{2026}

\title{Every Error has Its Magnitude: Asymmetric Mistake Severity Training for Multiclass Multiple Instance Learning}

\author{
Sungrae Hong$^1$ \hspace{3em} 
Jiwon Jeong$^1$ \hspace{3em} 
Jisu Shin$^1$ \hspace{3em}
Donghee Han$^1$ \\ 
Sol Lee$^1$ \hspace{3em}
Kyungeun Kim$^2$ \hspace{3em}
Mun Yong Yi$^1$\thanks{Corresponding Author}
\\
$^1$Korea Advanced Institute of Science and Technology, Daejeon, South Korea
\\
$^2$Seegene Medical Foundation, Seoul, South Korea
\\
$^1${\tt\small \{sr5043, zzioni, jisu3389, handonghee, leesol4553, munyi\}@kaist.ac.kr}
\\
$^2${\tt\small {kekim@mf.seegene.com}}
}

\begin{document}
\maketitle
\begin{abstract}
Multiple Instance Learning (MIL) has emerged as a promising paradigm for Whole Slide Image (WSI) diagnosis, offering effective learning with limited annotations. 
However, existing MIL frameworks overlook diagnostic priorities and fail to differentiate the severity of misclassifications in multiclass, leaving clinically critical errors unaddressed. We propose a mistake–severity–aware training strategy that organizes diagnostic classes into a hierarchical structure, with each level optimized using a severity-weighted cross-entropy loss that penalizes high-severity misclassifications more strongly. Additionally, hierarchical consistency is enforced through probabilistic alignment, a semantic feature remix applied to the instance bag to robustly train class priority and accommodate clinical cases involving multiple symptoms. An asymmetric Mikel’s Wheel-based metric is also introduced to quantify the severity of errors specific to medical fields. Experiments on challenging public and real-world in-house datasets demonstrate that our approach significantly mitigates critical errors in MIL diagnosis compared to existing methods. We present additional experimental results on natural domain data to demonstrate the generalizability of our proposed method beyond medical contexts.
\end{abstract}

\section{Introduction}
\begin{figure}[t!]
\centering
\includegraphics[width=\linewidth]{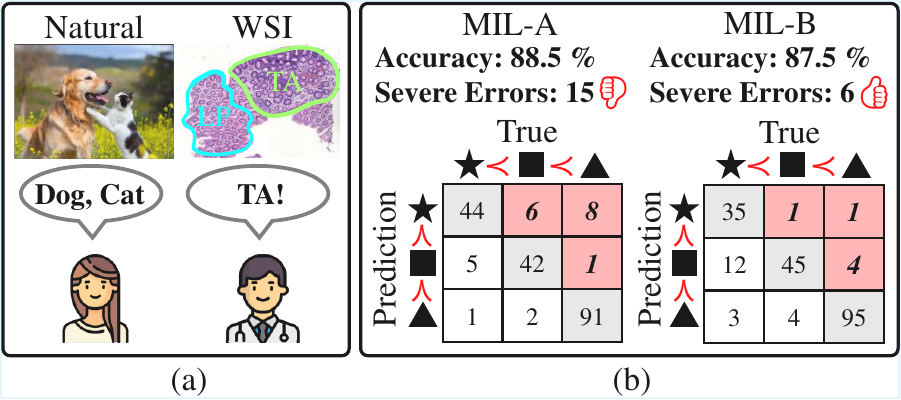}
\caption{The label characteristics that multiclass WSI have and the risks of applying MIL in the medical domain without accounting for mistake severity. \textbf{(a)} While every object in the natural domain is typically labeled, WSI only assigns the most urgent diagnosis among all observed complex findings. \textbf{(b)} Let \textcolor{red}{$\prec$} denotes more urgent diagnosis. Despite the model MIL-A being considered a traditionally "better" model than MIL-B because of high accuracy, it exhibits a higher number of severe misclassifications, which are shown in \crule[red!50!white!100]{0.25cm}{0.25cm} \textcolor{CarnationPink}{colored squares}. This comparison highlights that conventional MIL approaches can lead to unreliable assessments for medical applications.} \label{fig:fig1}
\end{figure}

Multiple Instance Learning (MIL) has revolutionized computational  pathology by allowing Whole Slide Image (WSI) based diagnosis to produce highly accurate results, fueled by advanced in deep learning (DL)~\cite{afonso2024multiple}. 

The primary purpose of MIL-based applications is to provide clinical assistance to pathologists, thus improving their overall judgment through initial diagnostic screening~\cite{quellec2012multiple}. For clinical adoption, MIL predictions must be reliably accurate while minimizing the risk of catastrophic misclassification-induced patient harm, underscoring that MIL systems must not only achieve strong overall accuracy but also explicitly suppress severe diagnostic errors (see Figure \ref{fig:fig1}).


Mistake severity (MS) is a research field that addresses the varying importance of classes in model predictions~\cite{wei2021fine}. The early structural embedding-centric approaches~\cite{deng2012hedging,bengio2010label} utilize a Word2Vec~\cite{mikolov2013efficient} pre-trained on Wikipedia~\cite{wikipedia} or incorporate the hierarchy using a distance-based regularizer in a Poincaré space~\cite{nickel2017poincare}. The dominant paradigm organizes multiclass into a hierarchical structure, introducing a classifier for every coarse-to-fine level~\cite{wei2021fine,hong2025priority}. Drawing on the foundation laid by Bertinetto et al.~\cite{bertinetto2020making}, which suggested using the Lowest Common Ancestor (LCA) for class hierarchy, these methods propose sequential learning~\cite{chang2021your}, explicit feature tuning~\cite{garg2022learning}, and class-wise weighted cross-entropy~\cite{polat2025class}.


Despite advances in MS, the solution that addresses the unique characteristics of clinical WSI has been largely neglected. As shown in Fig.~\ref{fig:fig1}(a), pathologists prioritize the most critical finding among co-existing symptoms in a multiclass WSI~\cite{brancati2022bracs,ginter2021histologic}, unlike natural images where every object is explicitly labeled. Thus, conventional approaches that assume priority only between distinct labels are ill-suited to WSIs, where implicit hierarchies exist within a single sample. Moreover, previous work defined the severity of the misclassification solely based on the distance between classes~\cite{polat2025class,zhao2024err}. This is problematic because medical risk depends on both directionality and distance. As illustrated in Fig.~\ref{fig:fig1}(b), misclassifying an urgent class as a trivial class poses a significantly more severe error than the opposite; hence, the clinical consequence of misdiagnosis is inherently asymmetric.

To overcome the limitations of MS on clinical WSI MIL, we propose \textbf{P}riority-\textbf{A}ware \textbf{M}istake \textbf{S}everity method (PAMS), an MIL-based method designed to address the asymmetry of misclassification issues. PAMS includes Mistake Severity Cross-Entropy (MSCE), a novel loss designed for multiclass clinical tasks characterized by asymmetric misclassification risk. Unlike naive cross-entropy, which only regularizes the predicted log-likelihood of the true class~\cite{hong2025priority,garg2022learning}, the MSCE applies a nuanced penalty by considering all possible severe classes the model may be confusing. In addition, recognizing that MIL needs to prioritize the most urgent diagnosis when complex findings are observed, similar to a pathologist's workflow, we propose Semantic Feature Remix (SFR). The SFR semantically remixes two samples with different labels, allowing MIL to explicitly learn the priority that exists among multiple classes. We also address a key limitation in existing MS metrics~\cite{zhao2024err}, which neglect the directionality of class priority. The proposed metrics quantify serious misdiagnosis by incorporating a distance-based penalty. Multiple experiments conducted using public and in-house multiclass datasets demonstrate that our proposed method effectively addresses the unique labeling characteristics of WSIs while showing significant strengths in terms of MS. We also present supplementary results on a natural domain dataset to further validate the generalizability of our approach.

We summarize our contributions as follows.
\begin{itemize}
\item{We address the critical, yet previously overlooked, multiclass WSI priority which is essential for the clinical MIL applications.}

\item{The proposed MSCE facilitates the prioritization of urgent diagnoses by severity-aware training. The SFR allows MIL to explicitly learn class priority by extracting semantic patches from a weakly labeled WSI and mixing them into a lower-severity WSI. Our proposed asymmetric metric quantifies mistake severity from a safety perspective.}

\item{We validate the proposed method using a challenging public dataset and a real-world dataset. Using these two, we highlight that existing MS methods often fail to adequately account for priorities in multiclass WSIs, establishing the validity of our approach. Additional experiment conducted on a natural domain dataset demonstrates the strong generalizability of our proposed method beyond clinical applications.}
\end{itemize}
\section{Related Work}
\subsection{Multiple Instance Learning}
Multiple Instance Learning (MIL) formulates a WSI $X_a$ as a bag of multiple patches $\{x_{a,k}\}_{k=1}^{n(X_a)}$, where $n(X_a)$ is the number of valid patches in $X_a$. In a binary classification, while the label for each patch, $y_{a,k}$, is unknown, the known WSI-level label $Y_a$ implies:
\begin{equation}
    Y_a = 
    \begin{cases}
    0 \text{ , iff } \sum_{k} y_{a,k} = 0 \\
    1 \text{ , otherwise.}
    \end{cases}
\end{equation}
In a multiclass setting, a WSI is assigned the label of the most severe diagnosis present, as determined by pathologists~\cite{brancati2022bracs, ginter2021histologic}. Given a set of $C$ classes $y_{a,k}\in\{c\}_{c=0}^{C-1}$, where a diagnosis $c+1$ is considered more urgent than $c$ (i.e., $0\prec\cdots\prec{C-1}$), the $Y_a$ is defined as follows:
\begin{equation}
    Y_a = \operatorname{max}_c\left(\{y_{a,k}\}_{k=1}^{n(X_a)}\right).
\end{equation}
A pre-trained feature extractor maps patches $\{x_{a,k}\}_{k=1}^{n(X_a)}$ into encoded instances $\{z_{a,k}\}_{k=1}^{n(X_a)}$, which a MIL aggregator then combines to predict the input $X_a$'s label, $\hat{Y}_a$.

Conventional MIL approaches identified critical features using mean, minimum, or maximum aggregators~\cite{ramon2000multi}. With the emergence of attention mechanisms, diagnostically key instances are now identified by learnable weighting layers~\cite{li2021dual}. AB-MIL~\cite{ilse2018attention} introduced gated attention to provide results with diagnostic explainability. Subsequent work~\cite{shao2021transmil, huang2023cross} leveraged a self-attention architecture~\cite{vaswani2017attention, xiong2021nystromformer} to overcome the limitations of prior methods by enabling information exchange between instances. DTFD-MIL~\cite{zhang2022dtfd} further extended this idea by constructing pseudo-bags to capture more nuanced inter-instance relationships. Graph-based architectures~\cite{godson2023multi, hou2022h} aggregate meaningful information by employing message passing. Similarly, HIPT~\cite{chen2022scaling} encodes the information from the finest patches into a class token for use at a coarser level.

Despite a decade of progress in MIL, most of the work has focused on binary classification and accuracy maximization, leaving multiclass settings largely understudied. Consequently, the community has not paid enough attention to diagnostic priority and the differing severity of errors in multiclass MIL.

\subsection{Mistake Severity}
Mistake severity (MS) aims to minimize decision errors that could lead to catastrophic outcomes in safety- and ethics-critical applications such as autonomous driving and medical diagnosis~\cite{garg2022learning,garfield2013measuring}. Deng et al.~\cite{deng2010does} demonstrated that leveraging semantic hierarchy information from WordNet~\cite{miller1995wordnet} can both improve classification performance and mitigate misclassification patterns. Bertinetto et al.~\cite{bertinetto2020making} proposed hierarchy-sensitive loss adaptations designed to reduce the hierarchical distance in top-$k$ predictions, but at the expense of top-1 accuracy. Chang et al.~\cite{chang2021your} mitigated the degradation of fine-grained accuracy caused by coarse class cross-entropy loss by partitioning the feature space to disentangle coarse and fine-grained features. Garg et al.~\cite{garg2022learning} introduced a loss and a regularizer designed to train a Hierarchy Aware Feature (HAF) space.

In real-world clinical settings, MIL must account for the severity of errors by prioritizing the reduction of errors on high-priority diagnoses—an essential requirement that remains largely underexplored.

\section{Method}
\subsection{Problem Formulation}
\begin{figure}[t!]
\centerline{\includegraphics[width=0.8\columnwidth]{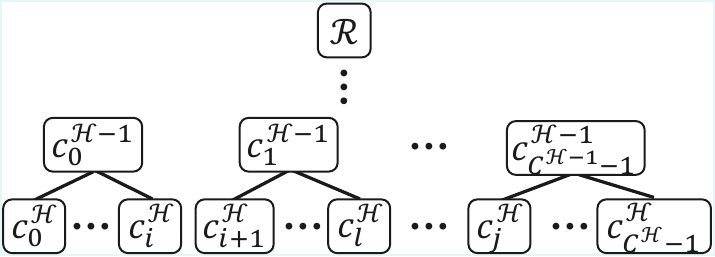}}
\caption{Structured class relationships from the finest hierarchy $\mathcal{H}$ to the root $\mathcal{R}$.}
\label{fig:hierarchy}
\end{figure}
The finest classes are structured into a hierarchy $h$, which is built by progressively grouping classes into coarser sets at each level, from $h=\mathcal{H}$ up to a single root $\mathcal{R}$ (i.e., $h=0$), as depicted in Fig.~\ref{fig:hierarchy}. For classes $\{c_n^{h-1}\}_{n=0}^{C^{h-1}-1}$ belonging to arbitrary hierarchy $h-1$, where $C^{h-1}$ is the number of classes in hierarchy $h-1$ s.t. $h\geq{2}$, if $c_i^{h-1}{\succ}c_j^{h-1}$ is met, then a $c^h_{i'}\succ{c^h_{j'}}$ is obvious for classes $c^h_{i'}\in{c^{h-1}_i}$ and $c^h_{j'}\in{c^{h-1}_j}$. At the same time, $c^{h}_{i'}{\equiv}c^h_{j'}$ can be for some classes $c^{h}_{i'}, c^h_{j'}\in{c^{h-1}_i}$.


\subsection{Mistake Severity Cross-Entropy Loss}
We introduce MIL classifiers, $f_\theta\leftarrow\bigcup_{h=1}^{\mathcal{H}}{f}_{\theta_h}$, for each hierarchy level $h$. This is to refine predictions at each level, given the structural relationships among the multiclasses. Existing approaches applied cross-entropy loss to each hierarchy:
\begin{equation}
    \mathcal{L}_{CE}=-\sum_{h=1}^{\mathcal{H}}\sum_{c=0}^{C^h-1}\tilde{Y}^h[c]\log{\hat{p}^{h}[c]}
\end{equation}\label{eq:CE}
where $\tilde{Y}^h\in\mathbb{R}^{1\times{C^h}}$ is the one-hot vectorized target $Y^h$, $[c]$ denotes a $c$-th component of the vector, and $\hat{p}^h\in\mathbb{R}^{1\times{C^h}}$ is a predicted probability. However, in multiclass WSIs, urgency is not uniform between classes. Conventional cross-entropy, which treats all classes within the same hierarchy as equally important, is unable to train the model to consider this priority. Furthermore, weighted cross-entropy~\cite{ho2019real,polat2025class} has the limitation of lacking the ability to account for directionality. Therefore, we propose a novel mistake severity cross-entropy (MSCE) loss that is sensitive to the directionality of errors and accounts for class priorities. The regularization weight matrix $M^h=[M^h_{ij}]\in\mathbb{R}^{C^h\times{C^h}}$ in the hierarchy $h$ is pre-defined as Eq.~\ref{eq:regularization_matrix}, where $\alpha>1$ is a scale hyper-parameter.
\begin{equation}\label{eq:regularization_matrix}
    M^h_{ij}=\left\{\begin{matrix}
    \alpha|i-j|\text{, if }c^h_i\succ{c^h_j}
    \\
    \hspace{1.4em}1\hspace{1.4em}\text{, otherwise} 
\end{matrix}\right.
\end{equation}
Then, $\mathcal{L}_{MSCE}$ is defined as the following term:
\begin{equation}
    \mathcal{L}_{MSCE}=-\sum_{h=1}^{\mathcal{H}}\underbrace{\hat{p}^h}_{\text{Pred.}}\underbrace{M^h}_{\text{Weight}}\underbrace{(\tilde{Y}^h)^\top}_{\text{True}}\sum_{c=0}^{C^h-1}\tilde{Y}^h[c]\log{\hat{p}^h[c]},
\end{equation}\label{eq:MS}
where $\hat{p}^hM^h(\tilde{Y}^h)^\top\in\mathbb{R}$ serves as a directional regularization weight with respect to both the ground truth and the predicted probability.

\subsection{Probability Alignment between Hierarchies}
To account for the fact that each hierarchy model, $f_{\theta_h}$, makes predictions on the same sample, we propose a method to ensure consistent alignment across their predictions. Building on this motivation and inspired by the approach in \cite{garg2022learning,hong2025priority}, we regularize the prediction probabilities of two consecutive hierarchy levels to be aligned. The vertical hierarchy alignment using Jensen-Shannon Divergence (JS) and Kullback-Leibler Divergence (KL) is defined as follows:
\begin{equation}\label{eq:vertical}
    \begin{aligned}
    \mathcal{L}_{HA}=\sum_{h=1}^{\mathcal{H}-1}\text{JS}(\hat{p}^h||\dot{p}^{h+1})
    \\
    =\frac{1}{2}\sum_{h=1}^{\mathcal{H}-1}\left(\text{KL}(\hat{p}^h||m)+\text{KL}(\dot{p}^{h+1}||m)\right)
    \end{aligned}
\end{equation}
where $m=\frac{1}{2}\times(\hat{p}^{h}+\dot{p}^{h+1})\in\mathbb{R}^{C^h}$. Given the predicted $\hat{p}^{h+1}\in\mathbb{R}^{C^{h+1}}$ at hierarchy $h+1$, the aligned target probability $\dot{p}^{h+1}\in\mathbb{R}^{C^h}$ for level $h$ is defined as:
\begin{equation}
    \dot{p}^{h+1}[i]=\sum_{{\forall}j:\hspace{0.3em}c^{h+1}_j{\in}c_i^h}\hat{p}^{h+1}[j]
\end{equation}\label{eq:p_dot}

The entire hierarchy of models, $f_\theta$, is trained in an end-to-end fashion by optimizing the loss $\mathcal{L}=\lambda_1\mathcal{L}_{MSCE}+\lambda_2\mathcal{L}_{HA}$, where $\lambda_1$ and $\lambda_2$ are tuning parameters.

\subsection{Semantic Feature Remix for Class Priority}
\begin{figure}[t]
\centering
\includegraphics[width=\linewidth]{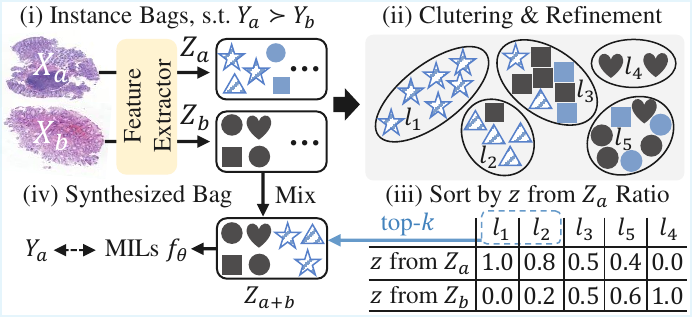}
\caption{
Conceptual Descriptions of SFR. Dashed figures represent unique symptoms in $Y_a$, where all figures are unlabeled and unknown. SFR extracts target cases and generates synthetic samples using only the available label $Y_a$.
}
\label{fig:sfr}
\end{figure}
MILs are required to prioritize the most urgent diagnosis when multiple pathologies are observed. However, the labeling convention for multiclass WSIs limits the acquisition or augmentation of labeled complex-finding samples~\cite{ginter2021histologic}. Leveraging recent findings that demonstrate the effectiveness of instance curation using weak semantic cues for model generalization~\cite{liu2024pseudo,shao2023lnpl}, we propose Semantic Feature Remix (SFR), which blends two instance bags with different priorities to simulate complex, multiclass cases.

Fig.~\ref{fig:sfr} illustrates the concept of the SFR. Preprocessing is conducted for $X_a$ and $X_b$ where $Y_a \succ Y_b$. Inspired by \cite{liu2024pseudo}, we cluster all instances from $Z_a\bigcup{Z_b}$, into $L$ clusters. Although the exact symptoms of $X_a$ are unknown, sorting $l \in L$ according to the patch ratio originating from $Z_a$ ensures that the distinct cases of $X_a$ occupy the majority.  As depicted in (ii) and (iii), these unique symptoms are primarily aggregated within the top-$k$ sorted elements of $l$. Finally, $Z_a$ instances from the top-$k$ clusters are added to $Z_b$ to form the synthesized semantic bag $Z_{a+b}$, which receives the higher priority label $Y_a$.

For clustering, we use the FAISS library~\cite{douze2024faiss,johnson2019billion}, which is implemented in low-level code and supports GPU parallel processing. We present the precise algorithm description and time complexity of SFR in the supplementary material.

\subsection{Asymmetric Mikel's Wheel Severity Metric}
\begin{figure}[!t]
\centerline{\includegraphics[width=0.8\columnwidth]{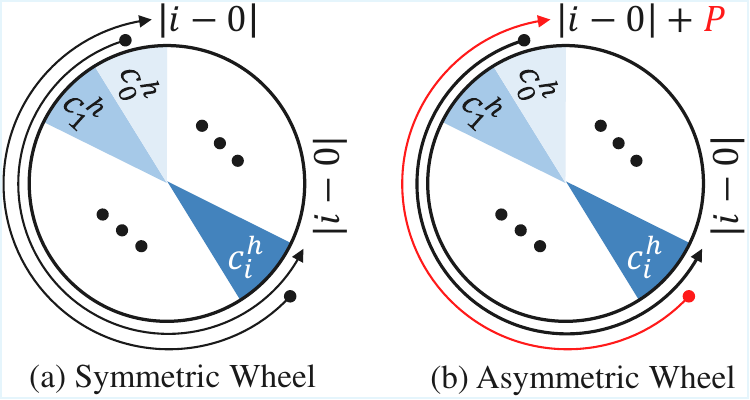}}
\caption{Two types of Mikel's Wheel, where $\bullet$ is the true class and $\blacktriangleright$ is the predicted class. (a) In a symmetric wheel, the distance between two classes $i$ and $j$ is the absolute difference. (b) Proposed asymmetric wheel imposes a penalty $P$ for misclassifying a high-priority true class.}
\label{fig:wheel}
\end{figure}
Quantifying the severity of a model's classification results is crucial in safety-critical tasks, especially in multiclass problems that involve a hierarchy of class priorities. Zhao et al.~\cite{zhao2024err} first addressed quantifying severity by introducing metrics that impose a greater penalty for misclassifications among higher-risk classes, with specific weights derived from Mikel's Wheel~\cite{mikels2005emotional,zhao2016predicting}. However, in the medical domain, this presents a fundamental drawback. For instance, misclassifying a malignant case as normal is an incomparably more severe error than the inverse, yet the symmetric Mikel's Wheel approach treats these identically. Therefore, we propose a severity metric based on an asymmetric Mikel's Wheel.

For prediction $\hat{Y}^h=c^h_i$ and true label $Y^h=c^h_j$ of a target hierarchy $h$, the distance between $c^h_i$ and $c^h_j$ is defined as $|{i-j}|$ (Fig.~\ref{fig:wheel}(a)). To address more severe misclassifications, we define a confusion weight that imposes an additional penalty $P\geq{1}$ when $c_i^h\succ{c^h_j}$, as shown in Fig.~\ref{fig:wheel}(b):
\begin{equation}\label{eq:weight}
    W^h_{i,j}=1+\mid{i-j}\mid+\space\mathbb{1}(c
    ^h_i{\succ}c^h_j)\times{P},\hspace{0.3em}\forall{W_{i,j}^h\geq{1}}
\end{equation}
Note that $W_{i,i}=1$ and that the matrix $W=[W_{i,j}]$ is asymmetric, such that $W^h_{i,j}\neq{W}^h_{j,i}$ for $c^h_i\not\equiv{c^h_j}$. Asymmetric Classification Confidence (AsCC) and Asymmetric Misclassification Confidence (AsMC), a severity metric that accounts for differing class importance, are given by Eq.~\ref{eq:AsCC} and Eq. \ref{eq:AsMC}, which are derived from Zhao et al.~\cite{zhao2024err}. AsCC measures the severity of all samples, while AsMC quantifies the severity only for misclassified samples.
\begin{equation}\label{eq:AsCC}
    \text{AsCC}=\frac{1}{\sum_{i,j}S_{i,j}^h}{\sum_{i=0}^{C^h-1}\sum_{j=0}^{C^h-1}\left({S_{i,j}^h}\times\frac{1}{W^h_{i,j}}\right)}
\end{equation}
\begin{equation}\label{eq:AsMC}
    \text{AsMC}=\frac{1}{\sum_{i{\neq}j}S_{i,j}^h}\sum_{i=0}^{C^h-1}\sum_{j=0}^{C^h-1}\left(\mathbb{1}(i\neq{j})\times\frac{{S^h_{i,j}}}{W^h_{i,j}-1}\right)
\end{equation}
Here, $S^h_{i,j}$ is the number of samples from class $c^h_i$ that were predicted as class $c_j^h$, and we use $P$ as 2.

\section{Experiments}
\subsection{Data Description}

\begin{table*}[t]
\centering
\resizebox{\textwidth}{!}{%
\begin{tabular}{l|cccccccc|cccccccc}
\hline
 &
  \multicolumn{8}{c|}{BRACS~\cite{brancati2022bracs}} &
  \multicolumn{8}{c}{In-house} \\ \cline{2-17} 
 &
  \multicolumn{4}{c|}{TransMIL~\cite{shao2021transmil}} &
  \multicolumn{4}{c|}{DTFD-MIL-AFS~\cite{zhang2022dtfd}} &
  \multicolumn{4}{c|}{TransMIL~\cite{shao2021transmil}} &
  \multicolumn{4}{c}{DTFD-MIL-AFS~\cite{zhang2022dtfd}} \\ \cline{2-17} 
\multirow{-3}{*}{} &
  ACC &
  AUC &
  AsCC &
  \multicolumn{1}{c|}{AsMC} &
  ACC &
  AUC &
  AsCC &
  AsMC &
  ACC &
  AUC &
  AsCC &
  \multicolumn{1}{c|}{AsMC} &
  ACC &
  AUC &
  AsCC &
  AsMC \\ \hline
Cross Entropy (CE) &
  \begin{tabular}[c]{@{}c@{}}40.23\\ \small$\pm$5.54\end{tabular} &
  \begin{tabular}[c]{@{}c@{}}74.90\\ \small$\pm$3.26\end{tabular} &
  \begin{tabular}[c]{@{}c@{}}58.48\\ \small$\pm$4.12\end{tabular} &
  \multicolumn{1}{c|}{\begin{tabular}[c]{@{}c@{}}50.18\\ \small$\pm$2.99\end{tabular}} &
  \begin{tabular}[c]{@{}c@{}}45.98\\ \small$\pm$0.94\end{tabular} &
  \begin{tabular}[c]{@{}c@{}}\uline{82.03}\\ \small$\pm$1.92\end{tabular} &
  \begin{tabular}[c]{@{}c@{}}61.98\\ \small$\pm$1.43\end{tabular} &
  \begin{tabular}[c]{@{}c@{}}48.07\\ \small$\pm$3.54\end{tabular} &
  \begin{tabular}[c]{@{}c@{}}90.41\\ \small$\pm$0.81\end{tabular} &
  \begin{tabular}[c]{@{}c@{}}99.16\\ \small$\pm$0.01\end{tabular} &
  \begin{tabular}[c]{@{}c@{}}93.13\\ \small$\pm$0.79\end{tabular} &
  \multicolumn{1}{c|}{\begin{tabular}[c]{@{}c@{}}44.16\\ \small$\pm$5.92\end{tabular}} &
  \begin{tabular}[c]{@{}c@{}}91.45\\ \small$\pm$0.49\end{tabular} &
  \begin{tabular}[c]{@{}c@{}}99.33\\ \small$\pm$0.06\end{tabular} &
  \begin{tabular}[c]{@{}c@{}}93.71\\ \small$\pm$0.30\end{tabular} &
  \begin{tabular}[c]{@{}c@{}}39.10\\ \small$\pm$1.65\end{tabular} \\
\rowcolor[HTML]{EFEFEF} 
Weighted CE (2:3:5) &
  \begin{tabular}[c]{@{}c@{}}42.15\\ \small$\pm$4.82\end{tabular} &
  \begin{tabular}[c]{@{}c@{}}78.15\\ \small$\pm$3.58\end{tabular} &
  \begin{tabular}[c]{@{}c@{}}59.22\\ \small$\pm$3.68\end{tabular} &
  \multicolumn{1}{c|}{\cellcolor[HTML]{EFEFEF}\begin{tabular}[c]{@{}c@{}}47.90\\ \small$\pm$0.76\end{tabular}} &
  \begin{tabular}[c]{@{}c@{}}47.51\\ \small$\pm$1.95\end{tabular} &
  \begin{tabular}[c]{@{}c@{}}81.57\\ \small$\pm$1.63\end{tabular} &
  \begin{tabular}[c]{@{}c@{}}63.16\\ \small$\pm$2.50\end{tabular} &
  \begin{tabular}[c]{@{}c@{}}47.91\\ \small$\pm$4.78\end{tabular} &
  \begin{tabular}[c]{@{}c@{}}\uline{91.88}\\ \small$\pm$0.41\end{tabular} &
  \textbf{\begin{tabular}[c]{@{}c@{}}99.41\\ \small$\pm$0.04\end{tabular}} &
  \begin{tabular}[c]{@{}c@{}}\uline{94.27}\\ \small$\pm$0.41\end{tabular} &
  \multicolumn{1}{c|}{\cellcolor[HTML]{EFEFEF}\begin{tabular}[c]{@{}c@{}}46.32\\ \small$\pm$3.90\end{tabular}} &
  \begin{tabular}[c]{@{}c@{}}91.65\\ \small$\pm$1.23\end{tabular} &
  \begin{tabular}[c]{@{}c@{}}99.33\\ \small$\pm$0.09\end{tabular} &
  \begin{tabular}[c]{@{}c@{}}93.99\\ \small$\pm$0.91\end{tabular} &
  \begin{tabular}[c]{@{}c@{}}43.55\\ \small$\pm$2.52\end{tabular} \\
Weighted CE (1:2:7) &
  \begin{tabular}[c]{@{}c@{}}41.76\\ \small$\pm$1.43\end{tabular} &
  \begin{tabular}[c]{@{}c@{}}76.37\\ \small$\pm$0.90\end{tabular} &
  \begin{tabular}[c]{@{}c@{}}58.96\\ \small$\pm$1.96\end{tabular} &
  \multicolumn{1}{c|}{\begin{tabular}[c]{@{}c@{}}46.92\\ \small$\pm$3.85\end{tabular}} &
  \begin{tabular}[c]{@{}c@{}}45.21\\ \small$\pm$1.08\end{tabular} &
  \begin{tabular}[c]{@{}c@{}}80.23\\ \small$\pm$0.16\end{tabular} &
  \begin{tabular}[c]{@{}c@{}}62.77\\ \small$\pm$1.08\end{tabular} &
  \begin{tabular}[c]{@{}c@{}}\uline{52.69}\\ \small$\pm$3.34\end{tabular} &
  \begin{tabular}[c]{@{}c@{}}89.17\\ \small$\pm$0.85\end{tabular} &
  \begin{tabular}[c]{@{}c@{}}99.14\\ \small$\pm$0.08\end{tabular} &
  \begin{tabular}[c]{@{}c@{}}92.29\\ \small$\pm$0.61\end{tabular} &
  \multicolumn{1}{c|}{\begin{tabular}[c]{@{}c@{}}44.60\\ \small$\pm$0.52\end{tabular}} &
  \begin{tabular}[c]{@{}c@{}}91.57\\ \small$\pm$0.45\end{tabular} &
  \begin{tabular}[c]{@{}c@{}}99.34\\ \small$\pm$0.05\end{tabular} &
  \begin{tabular}[c]{@{}c@{}}93.95\\ \small$\pm$0.24\end{tabular} &
  \begin{tabular}[c]{@{}c@{}}42.99\\ \small$\pm$3.79\end{tabular} \\
\rowcolor[HTML]{EFEFEF} 
Chang et al.~\cite{chang2021your} &
  \begin{tabular}[c]{@{}c@{}}\uline{47.51}\\ \small$\pm$2.17\end{tabular} &
  \begin{tabular}[c]{@{}c@{}}79.48\\ \small$\pm$1.55\end{tabular} &
  \begin{tabular}[c]{@{}c@{}}\uline{63.98}\\ \small$\pm$1.81\end{tabular} &
  \multicolumn{1}{c|}{\cellcolor[HTML]{EFEFEF}\begin{tabular}[c]{@{}c@{}}\uline{51.02}\\ \small$\pm$1.66\end{tabular}} &
  \begin{tabular}[c]{@{}c@{}}45.21\\ \small$\pm$4.82\end{tabular} &
  \begin{tabular}[c]{@{}c@{}}81.80\\ \small$\pm$2.21\end{tabular} &
  \begin{tabular}[c]{@{}c@{}}61.31\\ \small$\pm$4.14\end{tabular} &
  \begin{tabular}[c]{@{}c@{}}47.77\\ \small$\pm$2.55\end{tabular} &
  \begin{tabular}[c]{@{}c@{}}91.42\\ \small$\pm$0.41\end{tabular} &
  \begin{tabular}[c]{@{}c@{}}99.27\\ \small$\pm$0.02\end{tabular} &
  \begin{tabular}[c]{@{}c@{}}94.01\\ \small$\pm$0.06\end{tabular} &
  \multicolumn{1}{c|}{\cellcolor[HTML]{EFEFEF}\begin{tabular}[c]{@{}c@{}}49.07\\ \small$\pm$1.14\end{tabular}} &
  \begin{tabular}[c]{@{}c@{}}91.07\\ \small$\pm$0.98\end{tabular} &
  \begin{tabular}[c]{@{}c@{}}99.29\\ \small$\pm$0.10\end{tabular} &
  \begin{tabular}[c]{@{}c@{}}93.61\\ \small$\pm$0.83\end{tabular} &
  \begin{tabular}[c]{@{}c@{}}44.67\\ \small$\pm$0.41\end{tabular} \\
HXE $(\alpha=0.1)$~\cite{bertinetto2020making} &
  \begin{tabular}[c]{@{}c@{}}44.06\\ \small$\pm$1.95\end{tabular} &
  \begin{tabular}[c]{@{}c@{}}78.78\\ \small$\pm$1.41\end{tabular} &
  \begin{tabular}[c]{@{}c@{}}60.84\\ \small$\pm$2.06\end{tabular} &
  \multicolumn{1}{c|}{\begin{tabular}[c]{@{}c@{}}49.22\\ \small$\pm$4.93\end{tabular}} &
  \begin{tabular}[c]{@{}c@{}}43.30\\ \small$\pm$3.35\end{tabular} &
  \begin{tabular}[c]{@{}c@{}}81.14\\ \small$\pm$2.04\end{tabular} &
  \begin{tabular}[c]{@{}c@{}}59.46\\ \small$\pm$2.47\end{tabular} &
  \begin{tabular}[c]{@{}c@{}}45.91\\ \small$\pm$4.08\end{tabular} &
  \begin{tabular}[c]{@{}c@{}}90.76\\ \small$\pm$0.38\end{tabular} &
  \begin{tabular}[c]{@{}c@{}}99.21\\ \small$\pm$0.09\end{tabular} &
  \begin{tabular}[c]{@{}c@{}}93.63\\ \small$\pm$0.33\end{tabular} &
  \multicolumn{1}{c|}{\begin{tabular}[c]{@{}c@{}}50.69\\ \small$\pm$3.00\end{tabular}} &
  \begin{tabular}[c]{@{}c@{}}91.53\\ \small$\pm$1.40\end{tabular} &
  \begin{tabular}[c]{@{}c@{}}99.35\\ \small$\pm$0.16\end{tabular} &
  \begin{tabular}[c]{@{}c@{}}93.86\\ \small$\pm$1.08\end{tabular} &
  \begin{tabular}[c]{@{}c@{}}42.54\\ \small$\pm$1.62\end{tabular} \\
\rowcolor[HTML]{EFEFEF} 
HXE $(\alpha=0.5)$~\cite{bertinetto2020making} &
  \begin{tabular}[c]{@{}c@{}}41.38\\ \small$\pm$1.63\end{tabular} &
  \begin{tabular}[c]{@{}c@{}}79.52\\ \small$\pm$1.91\end{tabular} &
  \begin{tabular}[c]{@{}c@{}}58.18\\ \small$\pm$2.32\end{tabular} &
  \multicolumn{1}{c|}{\cellcolor[HTML]{EFEFEF}\begin{tabular}[c]{@{}c@{}}46.35\\ \small$\pm$4.59\end{tabular}} &
  \begin{tabular}[c]{@{}c@{}}47.51\\ \small$\pm$3.30\end{tabular} &
  \begin{tabular}[c]{@{}c@{}}81.79\\ \small$\pm$0.39\end{tabular} &
  \begin{tabular}[c]{@{}c@{}}62.77\\ \small$\pm$2.89\end{tabular} &
  \begin{tabular}[c]{@{}c@{}}47.32\\ \small$\pm$3.68\end{tabular} &
  \begin{tabular}[c]{@{}c@{}}90.99\\ \small$\pm$0.88\end{tabular} &
  \begin{tabular}[c]{@{}c@{}}99.19\\ \small$\pm$0.09\end{tabular} &
  \begin{tabular}[c]{@{}c@{}}93.78\\ \small$\pm$0.62\end{tabular} &
  \multicolumn{1}{c|}{\cellcolor[HTML]{EFEFEF}\begin{tabular}[c]{@{}c@{}}50.96\\ \small$\pm$2.79\end{tabular}} &
  \begin{tabular}[c]{@{}c@{}}91.03\\ \small$\pm$1.23\end{tabular} &
  \begin{tabular}[c]{@{}c@{}}99.31\\ \small$\pm$0.09\end{tabular} &
  \begin{tabular}[c]{@{}c@{}}93.44\\ \small$\pm$0.91\end{tabular} &
  \begin{tabular}[c]{@{}c@{}}40.87\\ \small$\pm$1.76\end{tabular} \\
Soft Labels $(\beta=5)$~\cite{bertinetto2020making} &
  \begin{tabular}[c]{@{}c@{}}45.21\\ \small$\pm$3.02\end{tabular} &
  \begin{tabular}[c]{@{}c@{}}74.18\\ \small$\pm$1.89\end{tabular} &
  \begin{tabular}[c]{@{}c@{}}61.67\\ \small$\pm$1.50\end{tabular} &
  \multicolumn{1}{c|}{\begin{tabular}[c]{@{}c@{}}48.53\\ \small$\pm$3.13\end{tabular}} &
  \begin{tabular}[c]{@{}c@{}}\uline{47.89}\\ \small$\pm$0.54\end{tabular} &
  \begin{tabular}[c]{@{}c@{}}80.80\\ \small$\pm$3.01\end{tabular} &
  \begin{tabular}[c]{@{}c@{}}\uline{63.83}\\ \small$\pm$0.06\end{tabular} &
  \begin{tabular}[c]{@{}c@{}}50.28\\ \small$\pm$1.87\end{tabular} &
  \begin{tabular}[c]{@{}c@{}}91.72\\ \small$\pm$0.24\end{tabular} &
  \begin{tabular}[c]{@{}c@{}}98.71\\ \small$\pm$0.21\end{tabular} &
  \begin{tabular}[c]{@{}c@{}}94.19\\ \small$\pm$0.20\end{tabular} &
  \multicolumn{1}{c|}{\begin{tabular}[c]{@{}c@{}}48.09\\ \small$\pm$3.24\end{tabular}} &
  \begin{tabular}[c]{@{}c@{}}91.76\\ \small$\pm$0.50\end{tabular} &
  \begin{tabular}[c]{@{}c@{}}98.79\\ \small$\pm$0.20\end{tabular} &
  \begin{tabular}[c]{@{}c@{}}94.07\\ \small$\pm$0.40\end{tabular} &
  \begin{tabular}[c]{@{}c@{}}43.16\\ \small$\pm$1.18\end{tabular} \\
\rowcolor[HTML]{EFEFEF} 
Soft Labels $(\beta=10)$~\cite{bertinetto2020making} &
  \begin{tabular}[c]{@{}c@{}}46.36\\ \small$\pm$1.43\end{tabular} &
  \begin{tabular}[c]{@{}c@{}}78.77\\ \small$\pm$1.53\end{tabular} &
  \begin{tabular}[c]{@{}c@{}}61.86\\ \small$\pm$2.01\end{tabular} &
  \multicolumn{1}{c|}{\cellcolor[HTML]{EFEFEF}\begin{tabular}[c]{@{}c@{}}46.35\\ \small$\pm$4.10\end{tabular}} &
  \begin{tabular}[c]{@{}c@{}}45.98\\ \small$\pm$0.03\end{tabular} &
  \begin{tabular}[c]{@{}c@{}}81.48\\ \small$\pm$1.52\end{tabular} &
  \begin{tabular}[c]{@{}c@{}}62.44\\ \small$\pm$0.82\end{tabular} &
  \begin{tabular}[c]{@{}c@{}}50.38\\ \small$\pm$3.52\end{tabular} &
  \begin{tabular}[c]{@{}c@{}}91.49\\ \small$\pm$0.54\end{tabular} &
  \begin{tabular}[c]{@{}c@{}}99.01\\ \small$\pm$0.11\end{tabular} &
  \begin{tabular}[c]{@{}c@{}}94.09\\ \small$\pm$0.41\end{tabular} &
  \multicolumn{1}{c|}{\cellcolor[HTML]{EFEFEF}\begin{tabular}[c]{@{}c@{}}49.82\\ \small$\pm$3.61\end{tabular}} &
  \begin{tabular}[c]{@{}c@{}}91.42\\ \small$\pm$1.02\end{tabular} &
  \begin{tabular}[c]{@{}c@{}}99.22\\ \small$\pm$0.08\end{tabular} &
  \begin{tabular}[c]{@{}c@{}}93.80\\ \small$\pm$0.78\end{tabular} &
  \begin{tabular}[c]{@{}c@{}}42.78\\ \small$\pm$2.18\end{tabular} \\
HAF~\cite{garg2022learning} &
  \begin{tabular}[c]{@{}c@{}}44.83\\ \small$\pm$0.94\end{tabular} &
  \begin{tabular}[c]{@{}c@{}}78.80\\ \small$\pm$2.51\end{tabular} &
  \begin{tabular}[c]{@{}c@{}}61.03\\ \small$\pm$1.78\end{tabular} &
  \multicolumn{1}{c|}{\begin{tabular}[c]{@{}c@{}}47.48\\ \small$\pm$4.55\end{tabular}} &
  \begin{tabular}[c]{@{}c@{}}46.36\\ \small$\pm$2.36\end{tabular} &
  \begin{tabular}[c]{@{}c@{}}81.39\\ \small$\pm$1.02\end{tabular} &
  \begin{tabular}[c]{@{}c@{}}62.46\\ \small$\pm$1.65\end{tabular} &
  \begin{tabular}[c]{@{}c@{}}49.16\\ \small$\pm$0.38\end{tabular} &
  \begin{tabular}[c]{@{}c@{}}91.18\\ \small$\pm$0.59\end{tabular} &
  \begin{tabular}[c]{@{}c@{}}99.28\\ \small$\pm$0.05\end{tabular} &
  \begin{tabular}[c]{@{}c@{}}93.87\\ \small$\pm$0.50\end{tabular} &
  \multicolumn{1}{c|}{\begin{tabular}[c]{@{}c@{}}49.69\\ \small$\pm$2.76\end{tabular}} &
  \begin{tabular}[c]{@{}c@{}}90.68\\ \small$\pm$1.38\end{tabular} &
  \begin{tabular}[c]{@{}c@{}}99.12\\ \small$\pm$0.02\end{tabular} &
  \begin{tabular}[c]{@{}c@{}}93.18\\ \small$\pm$1.11\end{tabular} &
  \begin{tabular}[c]{@{}c@{}}40.14\\ \small$\pm$3.06\end{tabular} \\
\rowcolor[HTML]{EFEFEF} 
Hong et al. $(\tau=10)$~\cite{hong2025priority} &
  \begin{tabular}[c]{@{}c@{}}47.13\\ \small$\pm$3.25\end{tabular} &
  \begin{tabular}[c]{@{}c@{}}\uline{79.80}\\ \small$\pm$1.29\end{tabular} &
  \begin{tabular}[c]{@{}c@{}}62.44\\ \small$\pm$1.97\end{tabular} &
  \multicolumn{1}{c|}{\cellcolor[HTML]{EFEFEF}\begin{tabular}[c]{@{}c@{}}45.54\\ \small$\pm$3.06\end{tabular}} &
  \begin{tabular}[c]{@{}c@{}}47.13\\ \small$\pm$1.45\end{tabular} &
  \begin{tabular}[c]{@{}c@{}}81.50\\ \small$\pm$1.07\end{tabular} &
  \begin{tabular}[c]{@{}c@{}}63.79\\ \small$\pm$1.32\end{tabular} &
  \begin{tabular}[c]{@{}c@{}}51.86\\ \small$\pm$1.81\end{tabular} &
  \begin{tabular}[c]{@{}c@{}}91.22\\ \small$\pm$1.34\end{tabular} &
  \begin{tabular}[c]{@{}c@{}}99.21\\ \small$\pm$0.16\end{tabular} &
  \begin{tabular}[c]{@{}c@{}}94.02\\ \small$\pm$0.93\end{tabular} &
  \multicolumn{1}{c|}{\cellcolor[HTML]{EFEFEF}\begin{tabular}[c]{@{}c@{}}\uline{53.10}\\ \small$\pm$1.76\end{tabular}} &
  \begin{tabular}[c]{@{}c@{}}91.11\\ \small$\pm$0.49\end{tabular} &
  \begin{tabular}[c]{@{}c@{}}99.35\\ \small$\pm$0.06\end{tabular} &
  \begin{tabular}[c]{@{}c@{}}93.80\\ \small$\pm$0.38\end{tabular} &
  \begin{tabular}[c]{@{}c@{}}49.16\\ \small$\pm$0.95\end{tabular} \\
Hong et al. $(\tau=15)$~\cite{hong2025priority} &
  \begin{tabular}[c]{@{}c@{}}42.53\\ \small$\pm$4.36\end{tabular} &
  \begin{tabular}[c]{@{}c@{}}77.68\\ \small$\pm$2.16\end{tabular} &
  \begin{tabular}[c]{@{}c@{}}60.23\\ \small$\pm$3.45\end{tabular} &
  \multicolumn{1}{c|}{\begin{tabular}[c]{@{}c@{}}50.20\\ \small$\pm$4.03\end{tabular}} &
  \begin{tabular}[c]{@{}c@{}}45.21\\ \small$\pm$2.87\end{tabular} &
  \begin{tabular}[c]{@{}c@{}}80.64\\ \small$\pm$1.60\end{tabular} &
  \begin{tabular}[c]{@{}c@{}}61.54\\ \small$\pm$2.92\end{tabular} &
  \begin{tabular}[c]{@{}c@{}}48.37\\ \small$\pm$3.77\end{tabular} &
  \begin{tabular}[c]{@{}c@{}}90.87\\ \small$\pm$2.04\end{tabular} &
  \begin{tabular}[c]{@{}c@{}}99.20\\ \small$\pm$0.36\end{tabular} &
  \begin{tabular}[c]{@{}c@{}}93.76\\ \small$\pm$1.48\end{tabular} &
  \multicolumn{1}{c|}{\begin{tabular}[c]{@{}c@{}}53.07\\ \small$\pm$4.85\end{tabular}} &
  \begin{tabular}[c]{@{}c@{}}\uline{91.69}\\ \small$\pm$0.67\end{tabular} &
  \begin{tabular}[c]{@{}c@{}}99.31\\ \small$\pm$0.07\end{tabular} &
  \begin{tabular}[c]{@{}c@{}}\uline{94.23}\\ \small$\pm$0.50\end{tabular} &
  \begin{tabular}[c]{@{}c@{}}\uline{50.59}\\ \small$\pm$4.30\end{tabular} \\
\rowcolor[HTML]{EFEFEF} 
CDW-CE~\cite{polat2025class} &
  \begin{tabular}[c]{@{}c@{}}44.83\\ \small$\pm$0.94\end{tabular} &
  \begin{tabular}[c]{@{}c@{}}79.06\\ \small$\pm$1.11\end{tabular} &
  \begin{tabular}[c]{@{}c@{}}61.05\\ \small$\pm$1.19\end{tabular} &
  \multicolumn{1}{c|}{\cellcolor[HTML]{EFEFEF}\begin{tabular}[c]{@{}c@{}}47.32\\ \small$\pm$2.16\end{tabular}} &
  \begin{tabular}[c]{@{}c@{}}\uline{47.89}\\ \small$\pm$3.30\end{tabular} &
  \begin{tabular}[c]{@{}c@{}}81.72\\ \small$\pm$1.05\end{tabular} &
  \begin{tabular}[c]{@{}c@{}}63.38\\ \small$\pm$3.51\end{tabular} &
  \begin{tabular}[c]{@{}c@{}}48.42\\ \small$\pm$2.23\end{tabular} &
  \begin{tabular}[c]{@{}c@{}}90.02\\ \small$\pm$0.16\end{tabular} &
  \begin{tabular}[c]{@{}c@{}}99.14\\ \small$\pm$0.18\end{tabular} &
  \begin{tabular}[c]{@{}c@{}}92.97\\ \small$\pm$0.23\end{tabular} &
  \multicolumn{1}{c|}{\cellcolor[HTML]{EFEFEF}\begin{tabular}[c]{@{}c@{}}47.14\\ \small$\pm$3.39\end{tabular}} &
  \begin{tabular}[c]{@{}c@{}}91.45\\ \small$\pm$1.22\end{tabular} &
  \textbf{\begin{tabular}[c]{@{}c@{}}99.37\\ \small$\pm$0.14\end{tabular}} &
  \begin{tabular}[c]{@{}c@{}}93.80\\ \small$\pm$0.91\end{tabular} &
  \begin{tabular}[c]{@{}c@{}}41.72\\ \small$\pm$1.32\end{tabular} \\ \hline
PAMS (Ours) &
  \textbf{\begin{tabular}[c]{@{}c@{}}47.59\\ \small$\pm$2.13\end{tabular}} &
  \textbf{\begin{tabular}[c]{@{}c@{}}80.61\\ \small$\pm$1.04\end{tabular}} &
  \textbf{\begin{tabular}[c]{@{}c@{}}64.92\\ \small$\pm$0.14\end{tabular}} &
  \multicolumn{1}{c|}{\textbf{\begin{tabular}[c]{@{}c@{}}55.65\\ \small$\pm$4.23\end{tabular}}} &
  \textbf{\begin{tabular}[c]{@{}c@{}}48.28\\ \small$\pm$1.92\end{tabular}} &
  \textbf{\begin{tabular}[c]{@{}c@{}}82.04\\ \small$\pm$0.73\end{tabular}} &
  \textbf{\begin{tabular}[c]{@{}c@{}}65.38\\ \small$\pm$0.83\end{tabular}} &
  \textbf{\begin{tabular}[c]{@{}c@{}}54.64\\ \small$\pm$3.36\end{tabular}} &
  \textbf{\begin{tabular}[c]{@{}c@{}}92.27\\ \small$\pm$0.47\end{tabular}} &
  \begin{tabular}[c]{@{}c@{}}\uline{99.31}\\ \small$\pm$0.03\end{tabular} &
  \textbf{\begin{tabular}[c]{@{}c@{}}94.84\\ \small$\pm$0.27\end{tabular}} &
  \multicolumn{1}{c|}{\textbf{\begin{tabular}[c]{@{}c@{}}56.20\\ \small$\pm$1.74\end{tabular}}} &
  \textbf{\begin{tabular}[c]{@{}c@{}}92.19\\ \small$\pm$0.20\end{tabular}} &
  \begin{tabular}[c]{@{}c@{}}\uline{99.36}\\ \small$\pm$0.06\end{tabular} &
  \textbf{\begin{tabular}[c]{@{}c@{}}94.66\\ \small$\pm$0.17\end{tabular}} &
  \textbf{\begin{tabular}[c]{@{}c@{}}51.78\\ \small$\pm$3.70\end{tabular}} \\ \hline
\end{tabular}%
}
\caption{Classification results on various mistake severity comparison methods and our proposed method. The best performance is denoted in bold, and the second best is underlined.}
\label{tab:main}
\end{table*}

\begin{figure}[!t]
\centerline{\includegraphics[width=0.9\columnwidth]{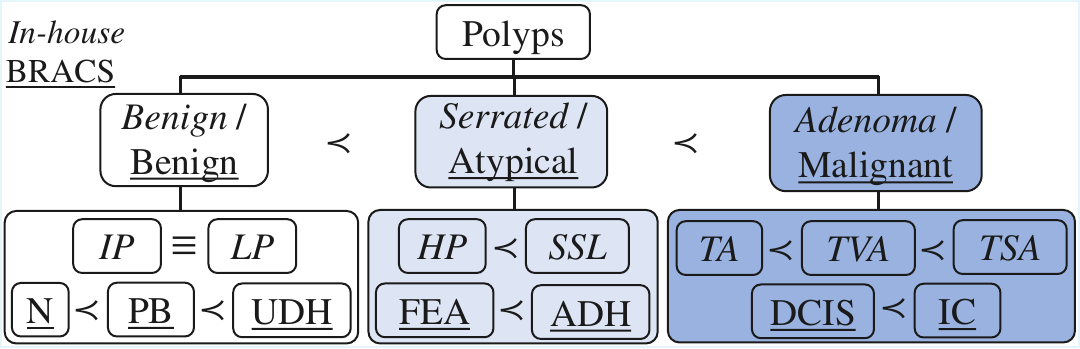}}
\caption{Data class hierarchy diagram of In-house and BRACS~\cite{brancati2022bracs} dataset. The In-house is presented in an italic font, whereas BRACS is underlined. We denote the class priority in each hierarchy using $\prec$ and $\equiv$.}
\label{fig:class}
\end{figure}

\noindent\textbf{BRACS}
The BReAst Carcinoma Subtyping (BRACS) dataset~\cite{brancati2022bracs} is a collection of hematoxylin and eosin (H\&E) stained histopathological images of breast carcinoma. The dataset, which contains 547 Whole-Slide Images (WSIs), is divided into seven classes: Normal (N), Pathological Benign (PB), Usual Ductal Hyperplasia (UDH), Flat Epithelial Atypia (FEA), Atypical Ductal Hyperplasia (ADH), Ductal Carcinoma in Situ (DCIS), and Invasive Carcinoma (IC). These are coarsely grouped into Malignant, Atypical, and Benign. The priority of these classes is illustrated in Fig.~\ref{fig:class}.

\noindent\textbf{In-house Dataset}
We used 4,734 WSIs from colon biopsy collected by Seegene Medical Foundation\footnote{This study was performed in line with the principles of the Declaration of Helsinki. Approval was granted by the Ethics Review Board SMF-IRB-2024-007 and KH2024-059.} from the real-world clinic. Expert pathologists annotated it at the WSI level, for which the most urgent diagnosis was designated. Among them, experts curated 182 challenging samples characterized by highly complex and mixed symptoms. The dataset consists of seven finest classes: Tubular Adenoma (TA), Tubulovillous Adenoma (TVA), Traditional Serrated Adenoma (TSA), Hyperplastic Polyp (HP), Sessile Serrated Lesion (SSL), Inflammatory Polyp (IP), and Lymphoid Polyp (LP). These are coarsely grouped into Adenoma, Serrated, and Benign, where their priority is depicted in Fig.~\ref{fig:class}.


The splits for both datasets are presented in the supplementary material. For the In-house dataset, 182 mixed symptom cases were used exclusively for testing. The remaining 4,552 cases were randomly partitioned. The BRACS dataset has a predefined split.

\subsection{Implementation Details}
\noindent\textbf{Comparison Methods}
To objectively validate our proposed method, we compare it with competing mistake severity methods. We employ cross-entropy (CE) and weighted CE, which uses manually assigned class weights, as our baselines. Chang et al.~\cite{chang2021your} proposed progressive learning across vertical hierarchies to build high-quality representations. Bertinetto et al.~\cite{bertinetto2020making} introduced soft labeling based on the lowest common ancestor (LCA) height in a given
hierarchy tree, along with a Hierarchical Cross-Entropy (HXE) term. Garg et al.~\cite{garg2022learning} introduced Hierarchy Aware Features (HAF), which regularizes model weights to capture class structure and enforce feature separation. Hong et al.~\cite{hong2025priority} proposed priority learning using a class hierarchy alignment loss and an implicit remix technique, which involves randomly sampling patches from the instance bag. The Class Distance Weighted Cross-Entropy (CDW-CE)~\cite{polat2025class} incorporates the class distance into the loss calculation.

\noindent\textbf{MIL Architectures}
We chose two MIL architectures as $f_{\theta_h}$ for each hierarchy. TransMIL~\cite{shao2021transmil} embeds instance relationships via self-attention~\cite{xiong2021nystromformer}, and its positional module merges local information using multi-sized convolutional operations. DTFD-MIL~\cite{zhang2022dtfd} performs a two-tier process by sampling the instance bag into multiple pseudo-bags. We adopted Aggregated Feature Selection (AFS) as the aggregator of DTFD-MIL.

\noindent\textbf{Experimental Settings}
We identified valid regions in WSIs using the Otsu algorithm~\cite{otsu1975threshold}, then extracted 256$\times$ patch sizes at 1.0 microns per pixel (MPP). All patches were mapped to a low-dimensional representation using a pre-trained feature extractor~\cite{kang2023benchmarking}. We trained the model using Adam optimizer~\cite{kingma2014adam} with $(\beta_1,\beta_2)=(0.9,0.999)$ and a learning rate of $1e-4$. We conducted experiments using an NVIDIA$^\circledR$ RTX A6000. We used parameters $(\alpha,\lambda_1,\lambda_2,L,T,k)=(1.6,2,1,11,6,6)$, where their sensitivity analysis is presented in the supplementary material.

\noindent\textbf{Evaluation Metrics}
For each MIL hierarchy $f_{\theta_h}$, we measure Accuracy (ACC) and Area Under the Curve (AUC). The proposed AsCC and AsMC are employed to measure the diagnostic importance of multiclass WSIs. The experiments are conducted using fixed seeds and the mean with standard deviation is reported.

\subsection{Classification Results and Mistake Severity}
\begin{figure}[!t]
\centerline{\includegraphics[width=\columnwidth]{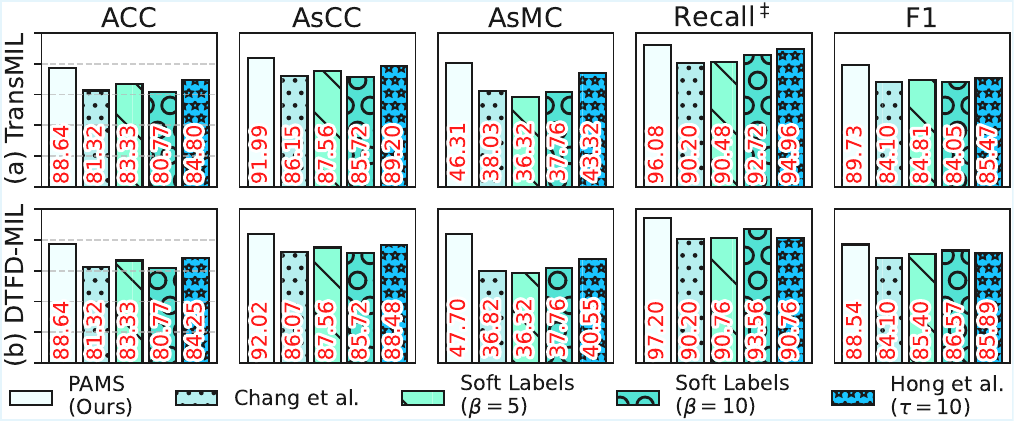}}
\caption{
Performance comparison on 182 mixed-symptom cases. Recall$^\ddagger$ is measured by designating Adenoma (i.e., TA, TVA, TSA) as the positive class.
}
\label{fig:radar}
\end{figure}

Table~\ref{tab:main} shows the quantitative performance of various MS methods. CE exhibits the lowest ACC, reflecting its struggle with classifying multiclass WSIs. Although weighted CE showed an improvement in ACC, it did not show a distinct advantage in severity metrics. Chang et al. achieved substantial growth in ACC and a corresponding growth in AsCC. However, the observed performance degradation within the DTFD-MIL architecture, coupled with a drop in AsMC, suggests that it is unlikely to serve as a robust solution for severity. Both HXE and Soft Labels (SL) improved ACC by formulating hierarchical relationships, but the gain in severity performance metrics was marginal. Similarly, HAF improved performance through feature hierarchy and contrastive-aware embedding, but its minimal or negative impact on DTFD-MIL demonstrates difficulty in generalization. Hong et al., using a random remix, achieved a drastic improvement on the In-house dataset, which contains complex symptoms. By demonstrating high AsMC across both model architectures, this work demonstrates the efficiency of the remix strategy for MS. However, it did not show a generalizable solution on the BRACS dataset. In contrast, PAMS demonstrates superior performance in all metrics. By recording the highest AsCC and AsMC against all compared methods, PAMS clearly validates its effectiveness for MIL training that explicitly prioritizes more severe classes.
\\
\noindent\textbf{Mixed-Symptom Cases}
Fig.~\ref{fig:radar} displays the test results of PAMS and the most competitive comparison methods on the 182 In-house mixed-symptom cases. PAMS achieves higher performance across all metrics, with the large performance gap in Recall$^\ddagger$, particularly indicating that PAMS effectively prioritizes the most urgent diagnosis in complex scenarios. Furthermore, the substantial difference in AsMC confirms that PAMS' error patterns carry a significantly lower MS risk.

\begin{table}[t]
\centering
\resizebox{0.9\columnwidth}{!}{%
\begin{tabular}{lcccc}
\hline
\multicolumn{5}{c}{TransMIL~\cite{shao2021transmil}}                                                                                                                                               \\ \hline
\multicolumn{1}{l|}{Ablation}                  & ACC                          & AUC                          & AsCC                         & AsMC                         \\ \hline
\multicolumn{1}{l|}{w.o. $\mathcal{L}_{MSCE}$} & 45.59\tiny{\color{red}-2.46} & 78.79\tiny{\color{red}-1.54} & 61.88\tiny{\color{red}-2.52} & 47.10\tiny{\color{red}-4.84} \\
\multicolumn{1}{l|}{w.o. $\mathcal{L}_{HA}$}   & 45.21\tiny{\color{red}-2.84} & 79.61\tiny{\color{red}-0.72} & 62.14\tiny{\color{red}-2.26} & 51.41\tiny{\color{red}-0.53} \\
\multicolumn{1}{l|}{w.o. SFR}                  & 47.51\tiny{\color{red}-0.54} & 76.88\tiny{\color{red}-3.45} & 62.99\tiny{\color{red}-1.41} & 47.92\tiny{\color{red}-4.02} \\
\multicolumn{1}{l|}{Ablated All}               & 40.23\tiny{\color{red}-7.82} & 74.90\tiny{\color{red}-5.43} & 58.48\tiny{\color{red}-5.92} & 50.18\tiny{\color{red}-1.76} \\ \hline
\multicolumn{5}{c}{DTFD-MIL-AFS~\cite{zhang2022dtfd}}                                                                                                                                           \\ \hline
\multicolumn{1}{l|}{w.o. $\mathcal{L}_{MSCE}$} & 43.68\tiny{\color{red}-3.19} & 80.70\tiny{\color{red}-1.34} & 62.19\tiny{\color{red}-3.19} & 54.45\tiny{\color{red}-0.19} \\
\multicolumn{1}{l|}{w.o. $\mathcal{L}_{HA}$}   & 44.83\tiny{\color{red}-3.45} & 81.82\tiny{\color{red}-0.22} & 61.96\tiny{\color{red}-3.42} & 51.31\tiny{\color{red}-3.33} \\
\multicolumn{1}{l|}{w.o. SFR}                  & 46.36\tiny{\color{red}-1.92} & 81.39\tiny{\color{red}-0.65} & 63.25\tiny{\color{red}-2.13} & 51.76\tiny{\color{red}-2.88} \\
\multicolumn{1}{l|}{Ablated All}               & 45.95\tiny{\color{red}-2.30} & 81.50\tiny{\color{red}-0.54} & 61.98\tiny{\color{red}-3.40} & 48.07\tiny{\color{red}-6.57} \\ \hline
\end{tabular}%
}
\caption{Ablation results for both models on the BRACS dataset. The red values denote the difference in metrics between the fully-equipped model and the ablated results.}
\label{tab:ablationBRACS}
\end{table}

\begin{figure*}[t!]
    \centering

    \footnotesize
    \begin{tabular*}{\textwidth}{cccc}
        \hspace{4em}\textbf{(a) ReMix}~\cite{yang2022remix} & \hspace{6.5em}\textbf{(b) PseudoMix}~\cite{liu2024pseudo} & \hspace{5.5em}\textbf{(c) RandomMix}~\cite{hong2025priority} & \hspace{6.4em}\textbf{(d) SFR (Ours)} \\
    \end{tabular*}

    \includegraphics[width=\textwidth]{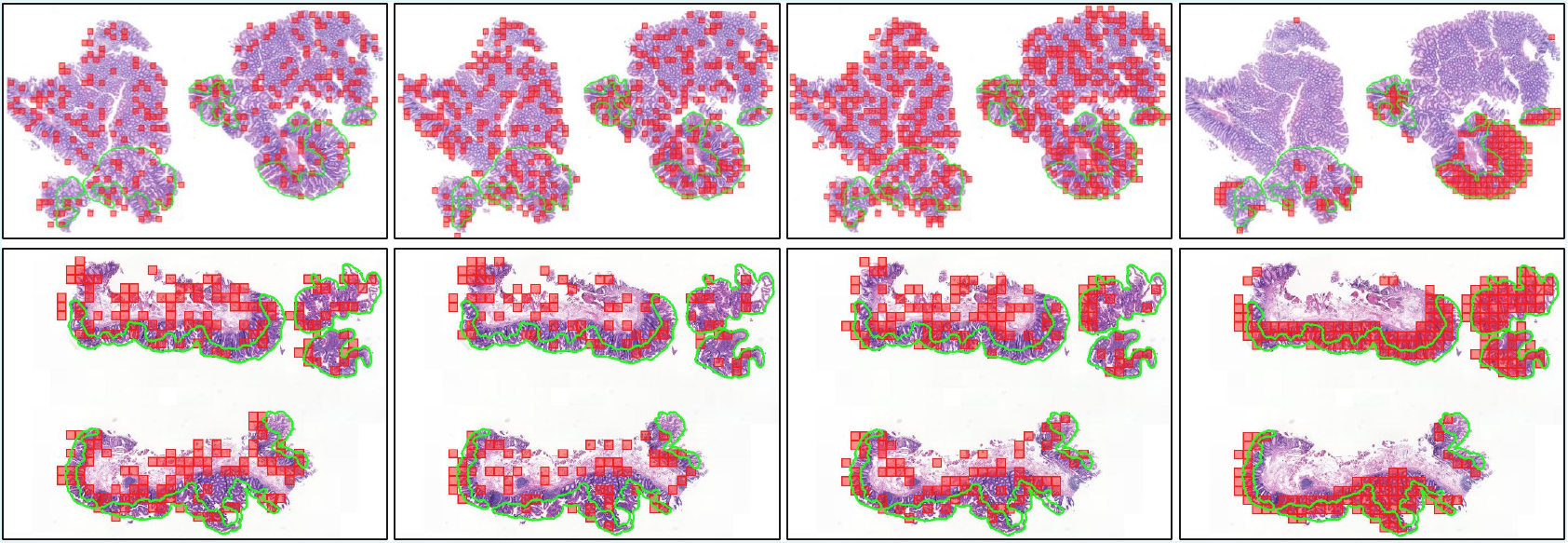} 
    \caption{
    Visualization of various instance remix strategies. \fcolorbox{green}{white}{\textbf{\color{green}The green polygon}} indicates the most severe diagnosis labeled by the experts. \textcolor{red}{Red boxes} \crule[red!100]{0.25cm}{0.25cm} highlight the patches selected by the remix approach to synthesize a new sample without pixel-level annotation.
    }
    \label{fig:remix_1}
\end{figure*}

\subsection{Ablation Study}
We conducted an ablation study on the BRACS dataset to quantitatively confirm the contribution of each PAMS component (Table~\ref{tab:ablationBRACS}). Ablation of $\mathcal{L}_{HA}$ resulted in a degradation of overall ACC and AUC, supporting the need to leverage hierarchical class alignment~\cite{chang2021your,bertinetto2020making}. Ablation of $\mathcal{L}_{MSCE}$ caused a severe decline in severity metrics, AsCC, and AsMC. Crucially, the large drop in AsMC confirms that $\mathcal{L}_{MSCE}$ is effective in suppressing critically severe mistakes in MIL. Ablating SFR led to a decrease in all metrics, which validates its role in providing robust generalization and sample robustness through data augmentation and semantic mixing. The fully ablated model yielded the poorest performance, demonstrating that the combination of the proposed components creates a highly effective synergy for MS training. We provide the ablation study on the In-house dataset in the supplementary material.

\begin{table}[t]
\resizebox{\columnwidth}{!}{%
\begin{tabular}{l|cc|cc}
\hline
\multirow{2}{*}{}                   & \multicolumn{2}{c|}{BRACS~\cite{brancati2022bracs}}                      & \multicolumn{2}{c}{In-house}              \\ \cline{2-5} 
                                    & ACC                    & AsMC                    & ACC                    & AsMC                    \\ \hline
-                                   & 40.23{\small$\pm$5.54} & 50.18{\small$\pm$2.29} & 90.41{\small$\pm$0.81} & 44.16{\small$\pm$5.92} \\
ReMix~\cite{yang2022remix}      & 40.61{\small$\pm$4.63} & 52.37{\small$\pm$4.45} & 90.84{\small$\pm$0.62} & 51.64{\small$\pm$1.66} \\
PseudoMix~\cite{liu2024pseudo}      & 43.30{\small$\pm$1.43} & 51.07{\small$\pm$0.40} & 89.25{\small$\pm$0.45} & 51.64{\small$\pm$2.99} \\
RandomMix~\cite{hong2025priority} & 44.06{\small$\pm$1.08} & \textbf{53.65{\small$\pm$4.45}} & 89.25{\small$\pm$1.75} & 54.34{\small$\pm$2.22} \\ \hline
SFR (Ours)                           & \textbf{45.59{\small$\pm$3.55}} & \uline{53.18{\small$\pm$1.78}} & \textbf{91.34{\small$\pm$0.09}} & \textbf{55.86{\small$\pm$0.52}} \\ \hline
\end{tabular}%
}
\caption{Various instance remix strategies and their performance.}
\label{tab:remix}
\end{table}

\subsection{Analysis on Semantic Feature Remix}
We compared the proposed SFR with various remix methods~\cite{yang2022remix,liu2024pseudo,hong2025priority} to verify that it generates semantical instance mixes. We used CE and TransMIL for the analysis.
\\
\noindent\textbf{Quantitative Comparison}
Every remix strategy improves MIL AsMC performance, validating its utility for MS in multicalss WSI (Table~\ref{tab:remix}). While ACC improved across all strategies on BRACS, the In-house dataset reported some degradation, suggesting accuracy gains are challenging when the room for performance improvement is limited. RandomMix shows high AsMC with computational cost efficiency. SFR helps classify complex cases, reduces fatal errors, and generally contributes to both ACC and AsMC over the comparison methods.
\\
\noindent\textbf{Remix Visualization}
Fig.~\ref{fig:remix_1} visualizes various WSI remix strategies. Since ReMix~\cite{yang2022remix} and PseudoMix~\cite{liu2024pseudo} are designed for prototyping, their remixing occurs uniformly in areas with unique WSI regions. Although these methods can augment samples by mixing global findings, they fail to selectively sample highly severe findings in the absence of pixel-wise annotations. Hong et al.~\cite{hong2025priority} mathematically proved that RandomMix includes at least one high-severity patch, but its random nature necessitates a large remix proportion to acquire a sufficient number of relevant patches for effective training. In contrast, SFR utilizes the semantic differences found between the two WSIs without explicit annotation to isolate and sort the higher-priority patches.

\begin{figure}[!t]
\centerline{\includegraphics[width=\columnwidth]{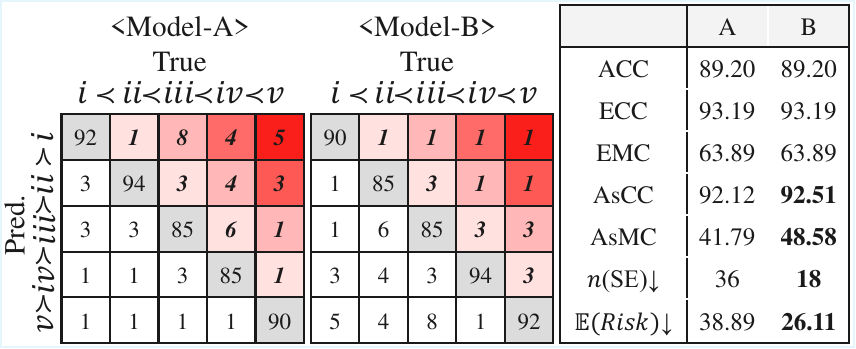}}
\caption{
The confusion matrix simulates the predictions of the two models, with severe errors highlighted by color. The more in-depth analysis and computation of $\mathbb{E}(\textit{Risk})$ are detailed in the supplementary material.
}
\label{fig:metric}
\end{figure}

\begin{table}[t]
\resizebox{\columnwidth}{!}{%
\begin{tabular}{l|cccc}
\hline
                                 & ACC                    & AUC                    & AsCC                   & AsMC                   \\ \hline
Cross Entropy (CE)               & 83.24{\small$\pm$0.20} & 98.57{\small$\pm$0.04} & 87.23{\small$\pm$0.18} & 34.84{\small$\pm$0.77} \\
Weighted CE (1:10)               & 73.69{\small$\pm$0.15} & 95.90{\small$\pm$0.16} & 79.79{\small$\pm$0.13} & 33.55{\small$\pm$0.21} \\
CO2~\cite{albuquerque2021ordinal}& 83.97{\small$\pm$0.18} & 98.55{\small$\pm$0.03} & 87.77{\small$\pm$0.15} & 34.73{\small$\pm$0.29} \\
CDW-CE~\cite{polat2025class}     & 84.11{\small$\pm$0.24} & 98.57{\small$\pm$0.01} & 87.87{\small$\pm$0.16} & 34.63{\small$\pm$0.40} \\ \hline
MSCE (Ours)                      & \textbf{85.64{\small$\pm$0.22}} & \textbf{98.80{\small$\pm$0.02}} & \textbf{89.12{\small$\pm$0.17}} & \textbf{35.70{\small$\pm$0.61}} \\ \hline
\end{tabular}%
}
\caption{CIFAR-10 performance on varietal entropy terms.}
\label{tab:cifar}
\end{table}

\begin{figure*}[t!]
    \includegraphics[width=\textwidth]{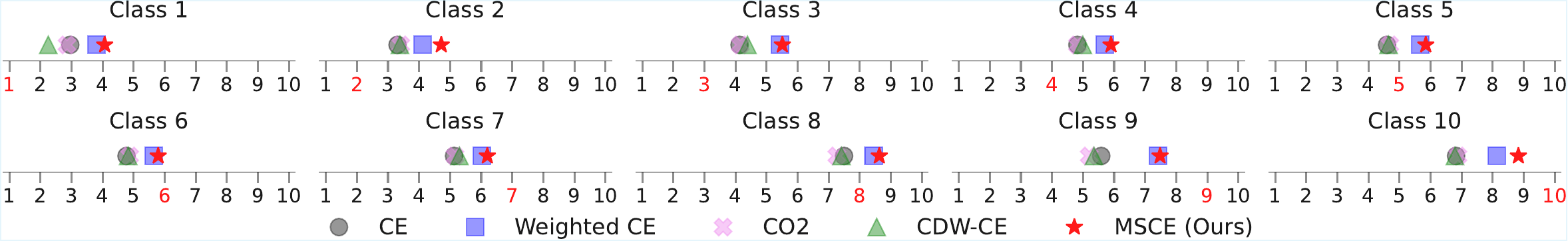} 
    \caption{
    Visualization of the expected class for various misclassified predictions. The values on the horizontal axis are the class index.
    }
    \label{fig:tsne}
\end{figure*}

\subsection{Mistake Severity Metric Analysis}
We present a dummy scenario in Fig.~\ref{fig:metric} to illustrate the practical utility of our proposed MS metrics. Models A and B share the same accuracy, which makes them indistinguishable by conventional means. Existing MS metrics such as ECC and EMC~\cite{zhao2024err} also produce identical scores even when a severity $i\prec\cdots\prec{v}$ is given. Because these metrics rely on a symmetric perspective, they fail to adequately prioritize Model-B, despite its significantly lower number of severe errors $n(\text{SE})$ and $\mathbb{E}(\textit{Risk})$. The proposed AsCC and AsMC, by assuming asymmetric severity, help identify the optimal model that is genuinely robust to fatal errors.

\subsection{Mistake Severity on Natural Data}
To validate the generalizability and scalability of the proposed MSCE, we present supplementary experiments where class priority is imposed on a natural image dataset which is consistent with prior studies~\cite{albuquerque2021ordinal,garg2022learning}. CIFAR-10~\cite{krizhevsky2009learning} classes are assigned severity in predefined index order\footnote{bird$\prec$car$\prec$cat$\prec$deer$\prec$dog$\prec$frog$\prec$horse$\prec$plane$\prec$ship$\prec$truck}. We use the ResNet-18~\cite{he2016deep} architecture for the analysis.
\\
\noindent\textbf{Quantitative Results}
Weighted CE performed worse than CE, showing difficulties in training in multiple classes, as depicted in Table~\ref{tab:cifar}. While CO2~\cite{albuquerque2021ordinal} and CDW-CE improved ACC and AsCC, their AsMC dropped relative to CE, confirming that increased accuracy is not intrinsically tied to reducing severe errors. MSCE performed the best on the severity metrics among the comparison methods. MSCE consistently produces the least severe error predictions by learning multiclass priority.
\\
\noindent\textbf{Error Prediction Severity}
We plotted the expected class of misclassified predictions to examine the severity of the error, which was calculated by $(1/|\mathcal{W}_c|)\times\Sigma_{i\in\mathcal{W}_c}\Sigma_{c=1}^{10}\{c\times\mathbf{\hat{p}}_i[c]\}\in[1,10]$, where $\mathcal{W}_c=\{i{\mid}(y_i=c){\wedge}(\hat{y}_i\neq{c})\}$. For the entire class, MSCE exhibited expected errors that leaned toward the most safe direction. In contrast, CE generally predicted errors in the risky direction. The results show that misclassifications can carry different levels of severity, depending on whether the model learns the class priority that determines error severity. Specifically, in the high-severity 6-10 classes, MSCE made expected predictions that were both closest to the ground truth and leaned towards the severe direction. Error predictions that distinguish a severe outcome while remaining close to the true label reflect the MS behavior required in safety-critical domains such as health and medicine.

\section{Conclusion}

In this paper, we introduce PAMS to address the distinctive labeling characteristics of clinical WSIs and to account for the often-overlooked asymmetric nature of misclassification risk in medical MIL. MSCE mitigates the limitations of conventional CE—which treats all errors uniformly—by applying severity-aware asymmetric penalties to every prediction pairing. SFR further semantically captures implicit class cues from WSIs, enabling the model to learn diagnostic priority with greater clarity. Our proposed metrics, AsCC and AsMC, surpass distance-based measures by precisely quantifying the asymmetric severity of misclassifications. Extensive experiments on public and in-house datasets show that PAMS substantially reduces severe diagnostic errors in MIL while effectively prioritizing more urgent classes. Supplementary results on natural image datasets further suggest that the proposed framework extends beyond medical imaging and is applicable to a wider range of safety-critical domains.

\section*{Acknowledgment} 
This work was supported by the National Research Foundation of Korea (NRF) grant funded by the Korean government (MSIT) under grant numbers RS-2022-NR068758. We are also deeply grateful for the generous support provided by the Seegene Medical Foundation in South Korea.

{
    \small
    \bibliographystyle{ieeenat_fullname}
    \bibliography{main}
}

\clearpage
\setcounter{page}{1}
\maketitlesupplementary


\section{Data Split}
\begin{table}[t]
\centering
\resizebox{\columnwidth}{!}{%
\begin{tabular}{ccccccccc}
\hline
\multicolumn{9}{c}{In-house}                                                                    \\ \hline
\multicolumn{1}{c|}{}           & TA      & TVA   & TSA    & HP    & SSL    & IP  & \multicolumn{1}{c|}{LP}  & $\sum$   \\ \hline
\multicolumn{1}{c|}{Train}      & 950     & 343   & 387    & 602   & 282    & 181  & \multicolumn{1}{c|}{445} & 3,190    \\
\multicolumn{1}{c|}{Validation} & 205      & 74    & 82     & 127    & 59     & 37  & \multicolumn{1}{c|}{94}  & 678      \\
\multicolumn{1}{c|}{Test}       & 300(95) & 81(6) & 102(18) & 136(8) & 114(55) & 38  & \multicolumn{1}{c|}{95}  & 866(182) \\ \hline
\multicolumn{9}{c}{BRACS~\cite{brancati2022bracs}}                                                                                               \\ \hline
\multicolumn{1}{c|}{}           & IC      & DCIS  & FEA    & ADH   & UDH    & PB  & \multicolumn{1}{c|}{N}   & $\sum$   \\ \hline
\multicolumn{1}{c|}{Train}      & 100     & 40    & 24     & 28    & 56     & 120 & \multicolumn{1}{c|}{27}  & 395      \\
\multicolumn{1}{c|}{Validation} & 12      & 9     & 6      & 8     & 9      & 11  & \multicolumn{1}{c|}{10}  & 65       \\
\multicolumn{1}{c|}{Test}       & 20      & 12    & 11     & 12    & 9      & 16  & \multicolumn{1}{c|}{7}   & 87       \\ \hline
\end{tabular}%
}
\caption{Data distribution over the classes. The values in parentheses represent 182 mixed cases, which were curated by the pathology experts.}
\label{tab:split}
\end{table}

Table~\ref{tab:split} summarizes the data splits for the In-house and BRACS datasets~\cite{brancati2022bracs}. For the In-house dataset, all 182 complex mixed-symptom cases, which were manually curated by experts, were reserved exclusively for the test set. The remaining 4,552 cases were randomly partitioned. Note that the 4,734 In-house samples are derived from a corresponding number of independent patients, ensuring no data leakage occurs. The BRACS dataset utilizes a predefined split.

\section{Details on Semantic Feature Remix}
\begin{algorithm}[!t]
\caption{Semantic Feature Remix} 
\label{alg:alg} 
\begin{algorithmic}[1] 
    \STATE \textbf{Input:} Instance bags $Z_a=\{z_{a,n}\}_{n=1}^{n(X_a)}$ and $Z_b=\{z_{b,n}\}_{n=1}^{n(X_b)}$, s.t. $Y_a\succ{Y_b}$, the number of clusters $L$, refinement iterations $T$, top-$k$ index $k({<}L)$
    \STATE {\textbf{Output:} Semantically remixed instance bag $Z_{a+b}$\\
            $//$ \textbf{1. Cluster Initialization}}
    \STATE {$p_{a+b}\leftarrow\left(1/{(|Z_a|+|Z_b|)}\right)\times(\Sigma_{n=1}^{|Z_a|}{z_{a,n}}+\Sigma^{|Z_b|}_{n=1}z_{b,n})$}
    \FOR{$z_i$ in $Z_a\bigcup{Z_b}$}
        \STATE{$s_i{\space\leftarrow\space}\text{sim}(p_{a+b},z_i)$ $//$ cosine similarity}
        \STATE{Allocate $z_i$ to $l$-th cluster $\varepsilon_l\in\mathcal{E}=\{\varepsilon_1,\cdots,\varepsilon_L\}$, 
               \\s.t. $s_i\in\left[-1+{2(l-1)}/{L},-1+{2l}/{L}\right]$}
    \ENDFOR {\\ $//$ \textbf{2. Cluster Refinement}}
    \FOR {$t=1$ to $T$}
        \FOR {$l=1$ to $L$}
            \STATE{$Z_l\leftarrow\{z|z\in\varepsilon_l\}$}
            \STATE{$p_l{\leftarrow}(1/|Z_l|)\times\Sigma_{z\in{Z_l}}z$}
        \ENDFOR
        \FOR {$z_i$ in $Z_a\bigcup{Z_b}$}
            \STATE{Allocate $z_i$ to $\varepsilon_l$, s.t. $l=\operatorname{argmin}_l{\text{sim}(p_l,z_i)}$}
        \ENDFOR
    \ENDFOR {\\$//$ \textbf{3. Remix}}
    \STATE {$\mathcal{E}'\leftarrow$ Sort clusters by highest $z$ proportion from $Z_a$}
    \STATE {$Z_{a+b}{\leftarrow}Z_b{\bigcup}\{z'_a|z'_a\in\mathcal{E}[1:k]\}$}
    \RETURN{$Z_{a+b}$}
\end{algorithmic} 
\end{algorithm}

Algorithm~\ref{alg:alg} illustrates the SFR process. (lines 3-7) Following the approach in \cite{liu2024pseudo}, we apply $L$-way clustering to the aggregated instances of $Z_a$ and $Z_b$. This process focuses on cases where $Y_a \succ Y_b$ to identify distinctive symptoms. (lines 8-16) $T$ refinements ensure the clustering of prototypical features. (lines 17-19) A cluster $l$ dominated by instances from $Z_a$ suggests that it captures typical features of the $Y_a$ condition, which are scarce in $Z_b$. Therefore, we sort the $L$ clusters by their $Z_a$ instance ratio. The synthesized bag $Z_{a+b}$ is formed by merging $Z_b$ with selected $Z_a$ instances from the top-$k$ clusters, with $Y_a$ serving as the definitive label.

\section{Sensitivity Analysis}
We present the parameter sensitivity analysis for both datasets, BRACS and In-house. All reported results were derived using the validation set, and TransMIL~\cite{shao2021transmil} was leveraged for the analysis.

\subsection{\texorpdfstring{$\lambda_1,\lambda_2$}{}, and \texorpdfstring{$\alpha$}{}}\label{section:7_1}
\begin{figure}[t!]
\centerline{\includegraphics[width=0.9\columnwidth]{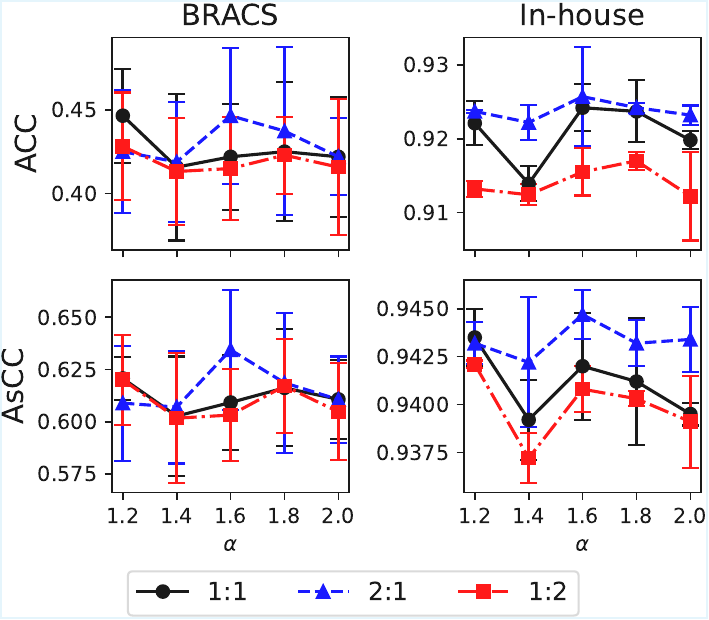}}
\caption{
Performance plot according to $\lambda_1,\lambda_2$, and $\alpha$.
}
\label{fig:sensi_lambda}
\end{figure}

Fig~\ref{fig:sensi_lambda} visualizes the validation performance with respect to the $\lambda_1,\lambda_2$ and variations in $\alpha$. The $(\lambda_1,\lambda_2) = (1,2)$ exhibited poor performance across both datasets. Conversely, the $(\lambda_1,\lambda_2) = (2,1)$  demonstrated strong overall performance, suggesting that applying a stronger penalty to MSCE is highly beneficial for WSI tasks that involve class priority. Specifically, optimal performance was consistently observed at $\alpha=1.6$.

\begin{figure}[t!]
\centerline{\includegraphics[width=0.9\columnwidth]{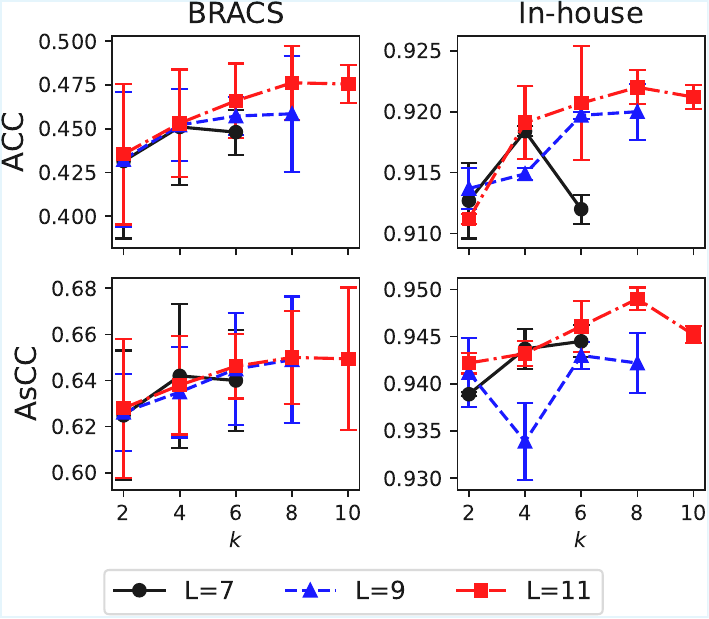}}
\caption{
Performance plot according to $L$ and $k$.
}
\label{fig:sensi_Lk}
\end{figure}

\subsection{Semantic Feature Remix}
Holding the parameters found in Section~\ref{section:7_1} constant, we visualize the performance sensitivity to the SFR parameters in Fig.~\ref{fig:sensi_Lk}. In general, selecting a value for $k$ that exceeded half of the total number of clusters (i.e., $k>L/2$) yielded acceptable performance across both datasets. Furthermore, increasing the number of clusters $L$ consistently provided a performance advantage over using the minimum number of clusters. This suggests that higher quality clusters can be obtained by increasing $L$, though at the expense of computational time.

\begin{figure}[t!]
\centerline{\includegraphics[width=0.95\columnwidth]{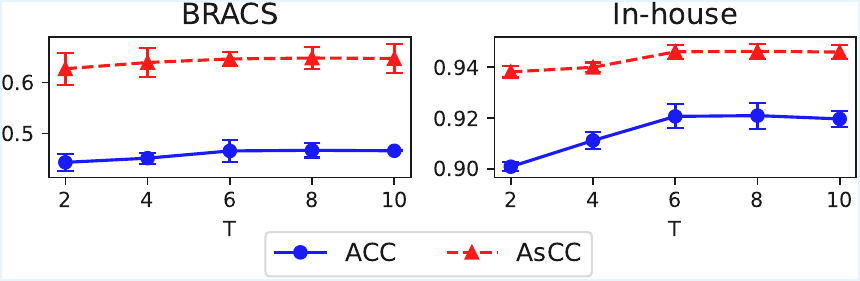}}
\caption{
Performance according to the refinement iterations $T$.
}
\label{fig:sensi_T}
\end{figure}

Fig.~\ref{fig:sensi_T} presents the performance of the two datasets as the cluster refinement iteration $T$ with $(L, k) = (11, 6)$. When $T$ is smaller than 6 (i.e., 2 and 4), both Acc and AsMC exhibit low performance across both datasets. Crucially, performance gains become negligible in both datasets when the $T$ exceeds 6. Consequently, we adopted the $T=6$ to ensure computational efficiency.

\section{Ablation Study on In-house Dataset}
\begin{figure}[!t]
\centerline{\includegraphics[width=0.9\columnwidth]{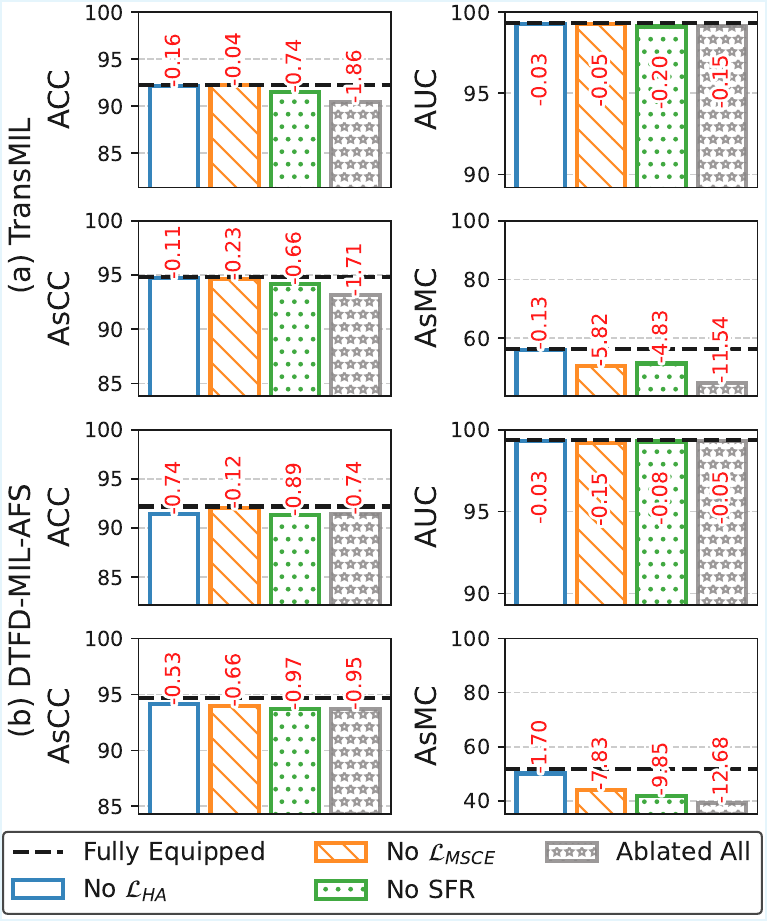}}
\caption{
Ablation results for both models on the In-house dataset. The red values denote the difference in metrics between the fully-equipped model and the ablated results.
}
\label{fig:ablation_in_house}
\end{figure}
We present the ablation study results for the In-house dataset in Fig.~\ref{fig:ablation_in_house}. In both model architectures, the ablation of $\mathcal{L}_{HA}$ resulted in a drop in ACC, underscoring the utility of leveraging a hierarchical approach in multiclass settings. Removing $\mathcal{L}_{MSCE}$ resulted in a more severe degradation of the severity metrics compared to removing $\mathcal{L}_{HA}$. We confirm that $\mathcal{L}_{MSCE}$ effectively mitigates severe errors in the inference stage by strongly regularizing high-severity errors during training. The SFR facilitated mistake severity ability in the MIL without imposing direct regulation on the model parameters. Its removal caused a significant drop in both AsCC and AsMC. The ablation of all components resulted in the largest performance decline, proving that the proposed PAMS is an effective solution for the multiclass WSI classification MIL task.


\section{Time Complexity Analysis on Remix}
\begin{table}[]
\resizebox{\columnwidth}{!}{%
\begin{tabular}{l|c|cc}
\hline
\multirow{2}{*}{Method} & \multirow{2}{*}{Time Complexity}        & \multicolumn{2}{c}{\begin{tabular}[c]{@{}c@{}}Mean Implementation\\ Time (s) / Sample\end{tabular}} \\ \cline{3-4} 
                        &                                         & BRACS~\cite{brancati2022bracs}                            & In-house                                           \\ \hline
ReMix~\cite{yang2022remix}       & $\mathcal{O}(L{\cdot}k{\cdot}T{\cdot}D{\cdot}n(X_i))$                        &    0.233                                         &0.061                                                    \\
PseudoMix~\cite{liu2024pseudo}   & $\mathcal{O}(L{\cdot}{k}{\cdot}T{\cdot}{n(X_i)})$ &    0.015                                         &0.004                                               \\
RandomMix~\cite{hong2025priority}& $\mathcal{O}(1)$                        &    0.005                                         &0.003                                               \\
SFR (Ours)                       & $\mathcal{O}(L{\cdot}{k}{\cdot}T{\cdot}{n(X_i})\log{n(X_i}))$       &    0.023                                         &0.005                                               \\ \hline
\end{tabular}%
}
\caption{
Time complexity of various remix methods and their practical implementation time on the training set of the two datasets.
}
\label{tab:time_complexity}
\end{table}

\begin{figure*}[t!]
    \centering
    \footnotesize
    \begin{tabular*}{\textwidth}{cccc}
        \hspace{4em}\textbf{(a) ReMix}~\cite{yang2022remix} & \hspace{6.5em}\textbf{(b) PseudoMix}~\cite{liu2024pseudo} & \hspace{5.5em}\textbf{(c) RandomMix}~\cite{hong2025priority} & \hspace{6.4em}\textbf{(d) SFR (Ours)} \\
    \end{tabular*}

    \includegraphics[width=\textwidth]{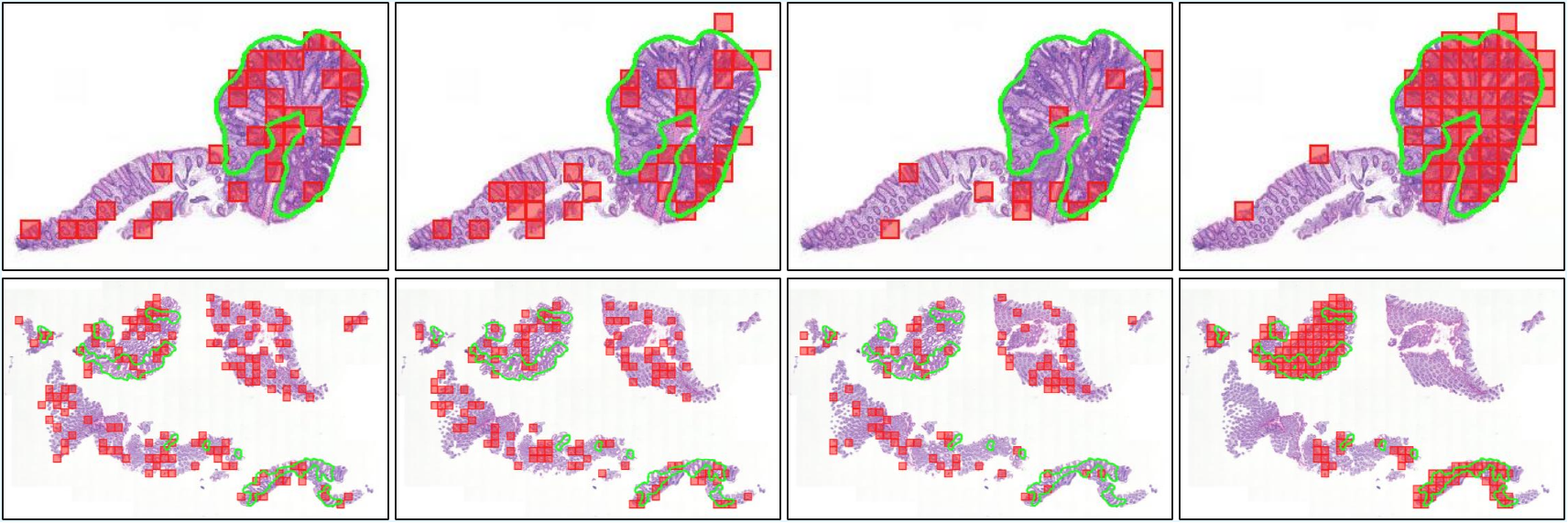} 
    \caption{
    Additional visualization results on various remix strategies. \fcolorbox{green}{white}{\textbf{\color{green}The green polygon}} indicates the most severe diagnosis labeled by the experts. \textcolor{red}{Red boxes} \crule[red!100]{0.25cm}{0.25cm} highlight the patches selected by the remix methodology to synthesize the sample without pixel-level annotation.
    }
    \label{fig:remix_supp}
\end{figure*}

\begin{figure*}[t!]
    \centering
    \footnotesize
    \begin{tabular*}{\textwidth}{cccc}
        \hspace{4em}\textbf{(a) ReMix}~\cite{yang2022remix} & \hspace{6.5em}\textbf{(b) PseudoMix}~\cite{liu2024pseudo} & \hspace{5.5em}\textbf{(c) RandomMix}~\cite{hong2025priority} & \hspace{6.4em}\textbf{(d) SFR (Ours)} \\
    \end{tabular*}

    \includegraphics[width=\textwidth]{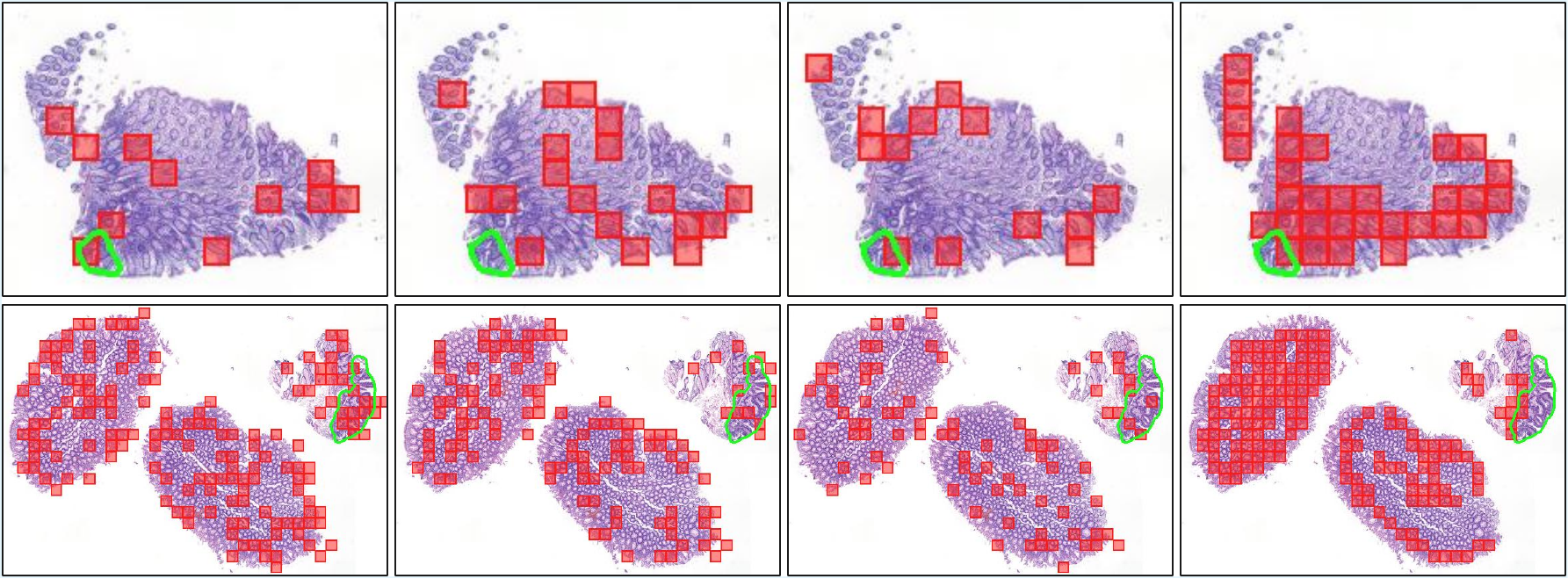} 
    \caption{
    Visualization of failure cases. \fcolorbox{green}{white}{\textbf{\color{green}The green polygon}} indicates the most severe diagnosis labeled by the experts. \textcolor{red}{Red boxes} \crule[red!100]{0.25cm}{0.25cm} highlight the patches selected by the remix methodology to synthesize the sample without pixel-level annotation.
    }
    \label{fig:remix_fail}
\end{figure*}

\begin{table*}[t]
\centering
\resizebox{0.8\textwidth}{!}{%
\begin{tabular}{l|ccc}
\hline
Method                            & Equation                                                                                                                                                                                                                                                                                              & Distance Penalty & Asymmetric Penalty \\ \hline
Cross Entropy (CE)                & $-\sum_c{\tilde{Y}[c]}\log{\hat{p}[c]}$                                                                                                                                                                                                                                                               & \xmark           & \xmark             \\
\rowcolor[HTML]{EFEFEF} 
Weighted CE                       & $-\sum_c{\tilde{Y}[c]}{w_c}\log{\hat{p}[c]}$                                                                                                                                                                                                                                                          & \xmark           & \xmark             \\
HXE~\cite{bertinetto2020making}   & \begin{tabular}[c]{@{}c@{}}$-\sum_{h}\exp{(-\alpha{h}(N^h))}\log{p(N^h|N^{h+1})}$\\ , where $h(N)\text{ is depth of node }N$\\ , $p(N^h|N^{h+1})=\frac{\sum_{A\in\text{Leaves}(N^h)}{p(A)}}{\sum_{B\in\text{Leaves}(N^{h+1})}{p(B)}}$\\ , and $p(N)=\prod_{h}p(N^h|N^{h+1})$\end{tabular}             & $\bigtriangleup$ & \xmark             \\
\rowcolor[HTML]{EFEFEF} 
CO2~\cite{albuquerque2021ordinal} & \begin{tabular}[c]{@{}c@{}}$-\sum_c{\tilde{Y}[c]}\log{\hat{p}[c]}+$\\ $\lambda\sum_c\mathbb{1}(c\geq{c'})\times\text{ReLU}(\delta+\hat{p}[c+1]-\hat{p}[c])+$\\ $\lambda\sum_c\mathbb{1}(c\leq{c'})\times\text{ReLU}(\delta+\hat{p}[c]-\hat{p}[c+1])$\\ , where $c'=\operatorname{argmax}_c\tilde{Y}$\end{tabular} & \cmark           & \xmark             \\
CDW-CE~\cite{polat2025class}      & \begin{tabular}[c]{@{}c@{}}$-\sum_c\left(\log(1-\hat{p}[c])\times|c-c'|^{\alpha}\right)$,\\ , where $c'=\operatorname{argmax}_c\tilde{Y}$\end{tabular}                                                                                                                                                & \cmark           & \xmark             \\
\rowcolor[HTML]{EFEFEF} 
MSCE (Ours)                       & $-\hat{p}{W}Y^{\top}\sum_c{\tilde{Y}[c]}\log{\hat{p}[c]}$                                                                                                                                                                                                                                             & \cmark           & \cmark             \\ \hline
\end{tabular}%
}
\caption{
The Cross Entropy and its associated severity loss terms. $\tilde{Y}\in\mathbb{R}^{1\times{C}}$ and $\hat{p}\in\mathbb{R}^{1{\times}C}$ indicate one-hot vectorized true label and predicted probability, respectively.
}
\label{tab:ce_terms}
\end{table*}

Table~\ref{tab:time_complexity} summarizes the time complexity and the actual implementation time per sample for the remix strategies. ReMix~\cite{yang2022remix} exhibits the largest time complexity, derived from the time required for prototyping plus the additional overhead $D$ for distributed instance generation. This results in an implementation time that exceeds the second-best method by a factor of 10 or more. PseudoMix~\cite{liu2024pseudo} requires only prototype clustering, giving it the smallest time complexity among computationally involved remix strategies. Its implementation time was shorter than both ReMix and SFR across both datasets. RandomMix~\cite{hong2025priority} has the shortest, constant time complexity because it relies only on random sampling for selection. SFR requires the time complexity for prototyping, with an added overhead for the sorting operation.

Despite the theoretical computational complexity of SFR and competing methods, practical implementation using FAISS~\cite{douze2024faiss,johnson2019billion} significantly minimized time constraints for their utilization. The empirical runtime of both SFR and PseudoMix was comparable to the $\mathcal{O}(1)$ complexity of RandomMix. Even ReMix, which has a significantly larger time complexity, yields an acceptable training time when executed on the datasets.

\section{Additional Qualitative Analysis on Remix}
We provide additional visualizations of various remix strategies in Fig~\ref{fig:remix_supp}. \textbf{(Top)} SFR successfully select patches that have symptoms, even in relatively narrow regions. By basing these semantically selected patches, SFR synthesizes a new instance bag containing only pertinent pathological characteristics. While ReMix and PseudoMix selected patches based on prototyping and clustering, this often resulted in the inclusion of some irrelevant patches. RandomMix provides the fastest execution time but may sometimes fail to capture pathological features entirely. \textbf{(Bottom)} SFR consistently selects higher-priority finding patches, even in complex-shaped tissue. Notably, SFR included even tiny areas, smaller than a single patch size, in the mixing. The absence of any sampling in the non-symptom tissue, which is located at top right side, demonstrates the precision of the pathological feature-based sorting and selecting. In contrast, comparison methods uniformly selected patches across all tissues, inevitably including numerous irrelevant patches.

Fig.~\ref{fig:remix_fail} presents failure cases. \textbf{(Top)} In instances where the pathology is extremely small and tissue presence is sparse, SFR selected patches irrelevant to the intended mixing source. This indicates that SFR struggles with too small regions of pathology because the clustering stage heavily incorporates low-level features, such as shape. Conversely, the comparison method PseudoMix completely failed to include a pathology patch. RandomMix did include pathology patches, but its probabilistic nature can make inclusion difficult when the lesion is small. \textbf{(Bottom)} SFR struggled to identify the correct features when the source WSI contained unique patterns specific to the finding but not representative of the tumor.

\section{Discussion on Mistake Severity Metrics}
\subsection{Risk Calculation}
\begin{equation}\label{eq:risk}
    \mathbb{E}(Risk)=\frac{1}{\sum_{i\neq{j}}S_{i,j}}\sum_i\sum_{j}\left(0.5^{\mathbb{1}(i{\succ}j)}\times{S_{i,j}}{W_{i,j}}\right)
\end{equation}
Equation~\ref{eq:risk} details the calculation of Risk used in Fig.~\ref{fig:metric}. We incorporate an increase in weight $2$ for severe errors and define $\mathbb{E}(\text{Risk})$ as the expected risk over the set of all misclassified samples.

\subsection{Accuracy, AsCC, and AsMC}
As Zhao et al.~\cite{zhao2024err} argued, severity metrics possess fundamentally different characteristics compared to conventional evaluation approaches. Accuracy is a metric that only counts samples that are classified correctly. While this can be interpreted as penalizing incorrect samples based on a perfect score $1$, it remains mathematically equivalent to simply counting the number of correct classifications. Consequently, most existing metrics exhibit a positive correlation with Accuracy (e.g., Recall and AUC). In contrast, AsMC has no relationship with Accuracy (i.e., the number of correct/incorrect samples is irrelevant). It is only related in the singular case where all test samples are correctly classified, resulting in $\text{AsMC}=\infty$. AsMC is determined solely by the severity pattern exhibited by the misclassified samples. Therefore, a model with high Accuracy is not guaranteed to achieve the high AsMC. The metric balancing these two perspectives is AsCC. AsCC can be approximated as the sum of Accuracy (representing the reward for correct samples) and AsMC (quantifying the appropriate severity of incorrect samples).

Consequently, a critical question emerges regarding the appropriate metric for model selection in safety-critical applications such as clinical diagnostics. The traditional approaches adopt the models with the highest Accuracy. However, MS studies have emphasized that in domains where the severity of incorrect predictions is critical, prioritization must shift to models exhibiting high AsMC. This suggestion requires a comprehensive selection standard that bridges accuracy and AsMC. We therefore recommend that model selection be guided by the AsCC metric when both overall Accuracy and mistake severity must be simultaneously optimized.



\section{Comparison on Entropy Terms}
We perform a comparative analysis of the standard Cross-Entropy (CE) term against several severity-aware CE variants (Table~\ref{tab:ce_terms}). CE only penalizes the prediction probability of the ground truth label and fails to account for severe error patterns and the model's overall prediction distribution across other classes. Weighted CE applies weighted regulation only to the true label index and neglects the global prediction pattern, similar to CE. HXE~\cite{bertinetto2020making} formulates classes into coarse-to-fine levels and implements hierarchical distance. This enforces a constraint s.t. the coarse prediction for a class node $N$ within a specific hierarchy $h$ matches the prediction probability of its fine-grained classes. However, the hierarchy enforced by HXE is fundamentally focused on probability alignment rather than class priority. CO2~\cite{albuquerque2021ordinal} defines an ordinal order among $C$ classes, which can be interpreted as severity. This approach, however, requires the strong assumption that all classes possess priority and cannot incorporate equivalence $\equiv$ relations. CDW-CE~\cite{polat2025class} is limited in that it uses only the distance between classes and treats the ground truth class as the most severe target. In contrast, the proposed MSCE can explicitly assign both dominance and equivalence relations for severity across all classes. Furthermore, it leverages the entire model prediction $\hat{p}$ to regulate the error pattern towards the less severe mistake.

\section{Limitations and Future Works}


\noindent\textbf{Contributions}
Our proposed PAMS framework introduces the crucial problem of MS to the MIL community, which previously focused solely on Accuracy, and highlights the risk of this to clinical deployment. We propose a novel remix method that robustly accounts for severity, specifically tailored to the unique labeling characteristics of multiclass WSIs. The two metrics we introduced, AsCC and AsMC, quantify the model’s mistake severity performance from a comprehensive view and solely from the perspective of misclassifications, respectively. By explicitly enforcing priority-aware training via MSCE, PAMS demonstrates superiority over existing severity comparison approaches across multiple metrics. Experiments conducted on challenging public and In-house datasets enable objective evaluation. Finally, we performed additional experiments on natural images to confirm the working mechanism and general scalability of MSCE.
\\
\noindent\textbf{Limitations}
During the course of this study, we identified several avenues for future work. Although MSCE enforces severity-aware training explicitly, it requires predefining the relationship between entire classes. This makes it difficult to assign precise severity values when class relationships are ambiguous or when the number of classes is extremely large. Furthermore, loss terms that involve the distance of the class can inherently lead to decreased training stability in tasks that involve numerous classes. This limitation, which is also shared by existing distance-based regularization MS approaches, requires a dedicated solution. Furthermore, we observed that MS research remains confined to multiclass classification settings. Although multi-label classification is a more generalized task, all existing MS approaches are currently specialized for multiclass problems.
\\
\noindent\textbf{Future Works}
To mitigate the identified limitations, we aim to develop a severity term that leverages both explicit and implicit signals. The objective is to autonomously discover patterns that are implicitly judged to be severe, while consistently respecting the explicitly defined severity. Furthermore, we seek to expand the scope from multiclass classification to multi-label classification. This is a more generalized task that is expected to contribute to the broader adoption of safety-centric deep learning ultimately.

\end{document}